\documentclass[smallextended]{svjour3} 
\usepackage{graphicx,epstopdf,multirow,multicol,booktabs,pgfplots}

\AtBeginDocument{%
  \paperwidth=\dimexpr
    1in + \oddsidemargin
    + \textwidth
    + 1in + \oddsidemargin
  \relax
  \paperheight=\dimexpr
    1in + \topmargin
    + \headheight + \headsep
    + \textheight
    + 1in + \topmargin
  \relax
  \usepackage[pass]{geometry}\relax
}

\usepackage{amsmath,bm}
\usepackage{amssymb,amsthm}
\usepackage[T1]{fontenc}
\usepackage[colorlinks,citecolor=blue]{hyperref}
\allowdisplaybreaks
\usepackage{setspace}
\usepackage{siunitx}
\usepackage{times}
\usepackage{siunitx}
\usepackage{pgfplots}

\usepackage{amssymb}
\usepackage{amsmath}
\usepackage{accents}

\usepackage{xcolor}

\usepackage{algorithmicx}
\usepackage{algpseudocode}
\usepackage{algorithm}

\pgfplotsset{compat=1.14}
\usepackage[backend=biber,style=ieee,doi=false]{biblatex}

\newcommand{\relu}{{\operatorname{ReLU}}}

\newcommand{\review}[1]{\textcolor{black}{#1}}

\AtBeginBibliography{\small}

\begin{document}
\title{Deployable, Data-Driven Unmanned Vehicle Navigation System in GPS-Denied, Feature-Deficient Environments}

\author{Sohum Misra$^{\dagger}$, Kaarthik Sundar$^{*}$, Rajnikant Sharma$^{\dagger}$, Kevin Brink$^{\ddagger}$
\thanks{$^{\dagger}$ Dept. of Aerospace Engg., University of Cincinnati, Ohio, USA. E-mail: \texttt{misra.sohum@gmail.com, sharmar7@ucmail.uc.edu}}\;
\thanks{$^{*}$Information Systems and Modeling, Los Alamos National
Laboratory, Los Alamos, New Mexico, USA. E-mail: \texttt{{kaarthik}@lanl.gov}}\;
\thanks{$^{\ddagger}$Air Force Research Laboratory, Eglin Air Force Base, Florida, USA. E-mail: \texttt{kevin.brink@us.af.mil}}\;
\thanks{The authors acknowledge Air Force Research Laboratory, Grant FA8651-16-1-0001 and LANL's Lab Directed Research and Development program (LDRD) project 20200016DR for funding this work.}\;
}

\maketitle
\begin{abstract}
This paper presents a novel data-driven navigation system to navigate an Unmanned Vehicle (UV) in GPS-denied, feature-deficient environments such as tunnels, or mines. The method utilizes landmarks that vehicle can deploy and measure range from  to enable localization as the vehicle traverses its pre-defined path through the tunnel. A key question that arises in such scenario is to estimate and reduce the number of landmarks that needs to be deployed for localization before the start of the mission, given some information about the environment. \review{The main focus is to keep the maximum position uncertainty at a desired value. In this article, we develop a novel vehicle navigation system in GPS-denied, feature-deficient environment by combining techniques from estimation, machine learning, and mixed-integer convex optimization. This article develops a novel, systematic method to perform localization and navigate the UV through the environment with minimum number of landmarks while maintaining desired localization accuracy. We also present extensive simulation experiments on different scenarios that corroborate the effectiveness of the proposed navigation system.}


\end{abstract}

\section{Introduction} \label{sec:Introduction}


The past decade has witnessed an accelerated growth in the application of  Unmanned Vehicles (UVs) owing to the versatility of UVs as a platform. The burgeoning research activities in fields such as controls, navigation, estimation, path planning, localization and many more with focus on UVs reflect its increased usage on a plethora of disparate applications viz. personal (photography) \cite{cheng2015aerial}, community (bridge inspection) \cite{chan2015towards}, business (package delivery) \cite{d2014guest}, military (surveillance, intelligence) \cite{hiltner2013drones, gregory2011view}, to name a few. Advancements made in the fields of computer science, material engineering, chip manufacturing, electrical, mechanical and aerospace engineering have enabled us to create better sensors, on-board processors, and UVs \cite{blais2004review, ScDaily2020, desai2016mos2, campbell2012review}. This in turn facilitated in developing complex algorithms to enable more precise and accurate navigation capabilities \cite{bristeau2011navigation, casbeer2005forest} for UVs. In particular, applications concerning search and rescue, environment mapping, payload delivery, etc. are based on autonomous navigation of the UVs through a specified path or sequence of waypoints. Most modern day autonomous navigation and path planning algorithms rely on a combination of Global Positioning System (GPS) and Inertial Measurement Units (IMUs) for accurate state (position, velocity, heading, etc.) estimation and localization to acquire seamless knowledge on position and heading of the UV with respect to its environment. However, these applications will be rendered ineffective in areas with unreliable GPS connectivity or in hostile territories where intentional GPS jamming is encountered \cite{Carroll2003,Hoey2005}. Furthermore, most indoor environments and many parts of terrain in an urban canyon of an outdoor environment do not have access to GPS; even if available, the access is intermittent and unreliable. Many authors have developed algorithms that utilize local features to aid navigation in GPS-denied or restricted areas. Simultaneous Localization and Mapping (SLAM) \cite{durrant2006simultaneous, bailey2006simultaneous, montemerlo2002fastslam}, especially Visual SLAM \cite{taketomi2017visual, yousif2015overview}, is widely used in such scenarios. However, in feature-deficient environments or places like dark tunnels or mines, even cameras may fail \cite{huang2017visual, howard2008real} to add significant information to enable localization or to keep the localization errors within desired uncertainty bounds. 

To address the shortcomings of Visual SLAM, several authors focused on solving the localization and mapping problem using RFIDs, \emph{i.e.}, range only beacons. D. Hahnel et. al. worked on improving the localization quality using RFID tags in \cite{hahnel2004mapping}. They placed 1000 RFID tags in random locations within a $25$ \si{\square\meter} area around the robot to meet their objective. A. Kleiner et. al. used RFID based technology to facilitate SLAM for search and rescue in \cite{kleiner2006rfid}. The RFID tags helped in creating a graph. The relative displacement between two tags were estimated and integrated with pose corrections from robot's odometry, IMU and laser-scans. However, the mapping itself was performed using Laser Range Finder. M. Beinhofer et. al. used a deployment strategy to artificially place range-based tags to improve data association for SLAM in \cite{beinhofer2013deploying}. In their approach, they use these tags to assist SLAM keeping the navigation tasks independent. This obviates the necessity of taking detours for landmark deployment. Moreover, they used a predetermined set of landmarks and their deployment relied on factors like remaining battery life, remaining number of landmarks on the robot, etc. Further, they assume that the environment contains features that can be mapped or used for localization. For example, during their experimental phase, they placed 70 RFID tags at randomly selected positions to serve as environmental features. V. A. Ziparo et. al. used RFID based exploration for multi-robot teams in \cite{ziparo2007rfid}. They have developed a task assignment method and have used RFID tags to facilitate exploration and multi-robot path planning alongside Laser Range Finder for mapping. Several other authors worked on improving SLAM efficiency using RFID tags \cite{kleiner2006rfid, vorst2008self}. However, all the works mentioned above assumes an environment containing features other than the tags that is primarily used for localization. All these aforementioned topics discusses strategies to improve mapping using RFID tags in a feature-rich environment. Moreover, they do not have a predetermined limitation on the optimal number of localization payloads (RFID tags) that they are required to carry to provide a good solution. Furthermore, these papers do not associate uncertainty bounds with vehicle localization estimates, neither do they consider how factors such as velocity, sensing range, noise, LOS would influence the localization efficiency. In this paper, we focus on optimally placing range-based beacons in an otherwise feature deficient environment to optimally deploy range sensors on-the-go and associate the position uncertainty bound or confidence factor along with the localization errors for the vehicle.

In this work, \review{we address the problem of localizing} and navigating the UV through closed indoor environments such as tunnels or mines using strategically deployed beacons (range sensors) at optimal locations in an otherwise, completely feature deficient environment. Any path for the UV inside a tunnel can be approximated as a set of straight lines (edges) joined via a series of waypoints (WPs). At any given point inside the tunnel, the cross-sectional can be approximated to be circular (having a radius) or rectangular (having a specific width). For the purpose of this article, we consider this environment to be dark and the environmental features to be repetitive. Therefore, it is safe to assume that algorithms like Visual SLAM can prove ineffective \cite{huang2017visual, howard2008real} in this scenario. Our objective is to navigate through such closed spaces while ensuring that the maximum uncertainty in the position estimate of the UV due to process and measurement noises during the WP-traversal lies close to a desired value. This desired bound in the maximum uncertainty in the position estimate is user-defined and it governs the traversal accuracy for a mission. However, it is mostly limited by the cross-section of the tunnel, i.e., the error in position estimates or the position uncertainty for the entire WP traversal cannot be more than the cross-section of the tunnel at most. To achieve this goal, we propose a data-driven approach to inject information in this otherwise information deficient environment borrowing techniques from neural networks, estimation, and mathematical programming. In particular, we assume that the UV can carry a limited number of range (RF) sensors, referred to as landmarks (LMs), and that it deploys these LMs as it traverses its trajectory. The vehicle is also equipped with a similar RF sensor like the Decawave UWB \cite{Decawave1001}that enables measurement of the vehicle's distance to a deployed landmark. The location where the LMs are deployed are estimated by the UV and we assume that each landmark's unique ID is known to the UV. The range measurements from the deployed landmarks in turn aid in position estimation. A simple approach to deploy LMs can be based of the instantaneous uncertainty in the position estimate. It should be noted that knowing the total length of the trajectory is sufficient to calculate the frequency of deploying these LMs, \emph{i.e.}, it is independent of WPs. We use WP-traversal as a proof-of-concept, but in actual scenario, a UV can rely upon range measurements from its surrounding and its estimated position and heading information to determine the direction it needs to travel. We assume that we have high confidence and accuracy regarding our starting location (depot) and have access to two initial landmarks near the start of the UV's path whose location will be exactly known (the reason for using exactly two initial landmarks will be made explicit in the later section). This is a fair assumption since information near the depot in most cases are readily available. Given this setup, this article presents algorithms to (i) compute where along the trajectory, the LMs should be deployed, (ii) compute the number of LMs the UV should carry, given various degrees of information about the topology of the tunnel, and (iii) estimate the position of the UV at every point in time using the deployed LMs as the vehicle traverses its path. All the algorithms are developed so that maximum position uncertainty is near a value that is specified a priori by the user and the errors in position estimates stay within the uncertainty bounds. In the next few paragraphs, we detail the related work of using deployable landmarks in GPS-denied and constrained environments. 

\review{The problem of localization and routing with limited information, especially in GPS-denied environment has been previously addressed in the literature \cite{icuas2017_gps,aiaa2018_mgps,dscc2018_gps,ol2019_gps}. In particular, authors in \cite{sharma2012graph} have developed a framework to localize a group of UVs cooperatively using bearing or range only measurements from objects (or landmarks) in the environment whose locations are known a priori. One key result from \cite{sharma2012graph} that we will put to effective use in this article is that as long as the UVs can share information among each other and each UV at least have path to two known LMs, the group of UVs can successfully localize themselves. In this context, we remark that for successful localization, the two known LMs and the UV should not be co-linear \cite{sharma2012graph}. Also, authors in \cite{misra2019single,acc2018_lpp} formulated and solved a optimization problem with sensor field-of-view constraint to perform WP traversal and LM placement given a starting location for the UV, a set of WPs to visit, and a set of potential LM placement locations. A feasible solution to the optimization problem would place LMs in a subset of potential locations that ensured that as the UV traverses its path, it can always obtain range measurements from at least two LMs so as to maintain observability \cite{sharma2012graph}. This article differs from the aforementioned work in the sense that here even the potential landmark locations are not known a priori i.e., but for the two landmarks in the start of the UV's path, all the other LM locations are only estimated and not known deterministically. Furthermore, this article provides a novel machine learning-based approach to identify landmark drop locations, deploy landmarks and in turn use range measurements from the deployed landmarks for localization of the vehicle. All of this is achieved while keeping the maximum uncertainty in the position estimate obtained by the estimation algorithm close to a desired value. In doing so, the observability guarantees do not carry over as the actual locations where the LMs are dropped are also estimated along with the vehicle's position and heading. Nevertheless, we show that by controlling the locations where the landmarks are dropped, the maximum uncertainty in the position estimate of the vehicle can still be kept under check.}

The problem of GPS-degraded relative navigation has been addressed previously by coupling Keyframing techniques with Visual SLAM \cite{leishman2013vision, leishman2014relative, ellingson2018relative}. While these approaches guarantee local observability, they still rely on visual odometry and being able to obtain information from the environment. Hence, in feature-deficient environments like the ones considered in this article, they tend to perform poorly. The same argument holds for many SLAM-based localization techniques \cite{durrant2006simultaneous, bailey2006simultaneous, montemerlo2002fastslam}.

In the next section, we present the formal problem statement and detail the novel research contributions. 

\section{Problem Setup and Statement} \label{sec:statement}
The following setup is considered throughout the rest of the article. We have a UV that needs to navigate through a tunnel. As it navigates through the tunnel, the UV drops RFID tags (landmarks) at regular time intervals to aid in localization. The landmarks provide range measurements to the UV which are used for localization. The UV needs to estimate its position and heading at all times such that the maximum uncertainty in the position estimate over the entire path-traversal is close to a user-specified value, $P_e$. The instantaneous uncertainty in the position estimate at time $t$ is defined as the trace of the square root of the position covariance matrix at time $t$ \cite{prentice2009belief} provided by the estimation algorithm; the formal definition of the instantaneous uncertainty in position estimate is detailed in the later sections. The goal of this article is to solve the aforementioned problem with three levels of information provided a priori: (i) we know that the tunnel is straight with a known length without any turns, (ii) we know the length of the tunnel and the number of turns, and (iii) we known full topology of the tunnel i.e., its length, the number of turns and the turn angle of each turn. 

For all the three scenarios, it is intuitively easy to see that uncertainty in position estimate decreases as the number of landmarks dropped increases i.e., dropping more landmarks before the previous landmarks are out of the vehicle's sensing range decreases this position uncertainty. On the other hand, the UV has a finite payload and  hence, this leads to the natural question of what is the optimal number of landmarks that the UV needs to carry and the location where they need to be dropped so that the maximum uncertainty position estimate (or simply maximum position uncertainty) is close to $P_e$, given the three levels of information; this question needs to be resolved before the start of the mission. In summary, our objectives are as follows:
\begin{enumerate}
    \item Compute the number of landmarks that the UV has to carry, and the locations where they need to be dropped along the vehicle's path to keep the maximum position uncertainty is close to $P_e$.
    \item Design an estimation algorithm for the UV to estimate its position and heading along its path and the drop locations of the landmarks using the range measurements provided by the landmarks it drops.
\end{enumerate}
\subsection{Assumptions} \label{eq:assumptions}
\begin{figure}
    \centering
    \includegraphics{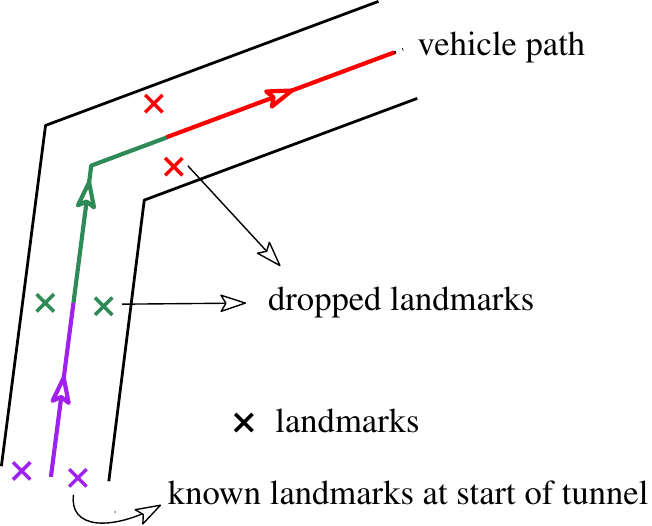}
    \caption{An illustration of a feasible solution. The color in the vehicle's path implies that the vehicle is receiving range measurements from the landmarks with the corresponding color.}
    \label{fig:illusration}
\end{figure}

The following assumptions about the tunnel, the vehicle, and drop locations are made throughout the rest of the article. Later in the Sec. \ref{sec:practical}, we comment on how to deal with a practical scenario when these assumptions are not necessarily valid. The first assumption we make is that the tunnel cross-section is either circular with a fixed radius or rectangular with a fixed width. Secondly, the UV is equipped with a mechanism to carry the landmarks and drop them in the floor of the tunnel as and when required. 
We characterize each drop location using the distance along the path from the start of the tunnel. At each drop location, we assume that two landmarks are dropped on the either side of the vehicle's path, symmetrically, at a fixed distance. We choose to drop two landmarks, one to the left of the path and another to the right. We do so because we know from the literature \cite{sharma2012graph} that, as long as the vehicle can receive range measurements from at least two landmarks whose position is known without uncertainty, it can perform localization. Hence, throughout the rest of the article, we assume that each landmark drop location is always associated with two landmarks being dropped. An illustration of the landmark drop and the vehicle's path through the tunnel is shown in Fig. \ref{fig:illusration}. 

\review{
\subsection{Overview of solution approach} \label{subsec:sol-approach}
In order to meet the objective of the problem, we break the problem into four sub-problems defined as follows. 
\begin{enumerate}
    \item The first sub-problem is that of state estimation. In this sub-problem, we develop a state estimation algorithm to estimate the position and heading of the vehicle as it traverses through its path assuming that we know the positions, along the path, where the landmarks will be dropped by the vehicle. In the process, we also identify all the input factors or features of the mission and the sensors that affect the position uncertainty. 
    \item The second sub-problem is focused on modeling the dependence between the identified input factors and the maximum position uncertainty using a Deep Neural Network (DNN). 
    \item The action of this DNN on the input factors is then equivalently reformulated as a Mixed Integer Linear Program (MILP). This MILP is then used to solve the inverse problem of finding the frequency at which the landmarks should be dropped so that the maximum position uncertainty is close to the desired value, $P_e$.
    \item The final sub-problem is aimed at adjusting the drop locations of the landmarks using information on the topology of the tunnel.
\end{enumerate}  
The presentation of the algorithms to solve the aforementioned sub-problems is followed by a discussion on points of failure of the overall system, novel methods to address this failure points, and a summary of the overall system architecture. }

\subsection{Paper organization} \label{subsec:paper-organization}
The remainder of this paper is organized as follows: in Sec. \ref{sec:state-estimation}, we present the system and measurement models \review{followed by the state estimation to estimate the position of the vehicle} and the dropped landmarks for a fixed landmark drop location. The Sec. \ref{sec:dnn} models the dependence between various input factors and the position uncertainty using an artificial neural network and presents an MILP reformulation of the DNN to solve the inverse problem of computing the drop locations given a desired maximum position uncertainty. Then, Sec. \ref{sec:drop-adjustment} presents algorithms to adjust the drop locations provided by the MILP to account for loss of line of sight in tunnels with turns. Finally the Sec. \ref{sec:arch} -- \ref{sec:conclusion} presents the details on other sensors used in the navigation systems, the system's workflow, simulation results, and concludes the article, respectively. 

\section{State Estimation Algorithm} \label{sec:state-estimation}
In this section, we present a state-estimation algorithm to estimate the position and heading of the vehicle when the information on the drop locations of the landmarks along the vehicle's path is known a priori. To that end, we first present some notations. At time $t=0$, the vehicle is at the start of the tunnel and it is assumed that there are two known landmarks $\ell_a$ and $\ell_b$ with known distinct locations $(x_a, y_a)$ and $(x_b, y_b)$, respectively, at the start of the tunnel. We let $L = \{\ell_1, \ell_2, \cdots, \ell_n\}$ denote the set of landmarks whose drop locations have been decided before the start of the mission. As the vehicle navigates through the tunnel, it drops the landmark $\ell_i$ at its corresponding drop location. The estimation algorithm, apart from estimating the position and heading of the vehicle $v$, also dynamically estimates the position of the landmarks that the vehicle drops during its mission. Next, we present a dynamic model for the vehicle $v$ and the landmarks in the set $L$.

\subsection{Dynamic models for the UV and the landmarks} \label{subsec:models}
For the vehicle, we use $x(t)$, $y(t)$ and $\psi(t)$ to denote the $x$-coordinate, $y$-coordinate, and the heading of the vehicle at time $t$. The state of the vehicle at any time $t$ is denoted by the vector $\bm x(t) = \left[ x(t),  y(t), \psi(t) \right]^{\top}$. The control variables for the vehicle is its speed $v$ and angular velocity $\omega(t)$. We let $\bm u = \left[ v, \omega(t) \right]^{\top}$ denote the control variable vector. Throughout the rest of the article, we do not show the explicit dependence of the state and control variables on time unless ambiguity arises. The following two-dimensional differential drive model is used throughout the rest of the article.
\begin{flalign}
\dot{\bm x} = 
\begin{bmatrix} \dot{x} \\ \dot{y} \\ \dot{\psi} \end{bmatrix} = 
\begin{bmatrix} v \cos \psi \\ v \sin \psi \\\omega \end{bmatrix} = \bm f(\bm x, \bm u) \quad \forall t \geqslant 0. \label{eq:uv-model}
\end{flalign}
In this context, we remark that it is trivial to extend this kinematic model to three dimensions and generalize the estimation algorithm to that setting. Nevertheless, we do not do so to keep the presentation simple.  

As for the landmarks, we let $\bm x_i = (x_i,  y_i)$ denote the drop location of the landmark $\ell_i$. We also let $t_i>0$ denote the time at which the landmark $\ell_i$ is dropped. Since, the landmarks are stationary, the dynamic model for the landmark $\ell_i$ is as follows:
\begin{flalign}
    \dot{\bm x}_i = \begin{bmatrix} \dot{x}_i(t) \\ \dot{y}_i(t) \end{bmatrix} = \begin{bmatrix} 0 \\ 0 \end{bmatrix} \quad \forall t \geqslant t_i. \label{eq:landmark-model}
\end{flalign}
The overall goal of the estimation algorithm is to estimate $\bm x$ and $\bm x_i$ for all $i \in \{1, \dots, n\}$. In the next section, we present the measurement model that is used to drive the estimation algorithm. 

\subsection{Measurement model}  \label{subsec:measurement}
As mentioned in the problem setup in Sec. \ref{sec:statement}, the vehicle is equipped with a range sensor using which it can receive range measurements from a landmark $\ell_i$; we let ${\rho}^{\max}$ denote the sensing range of the sensor. The vehicle can obtain range measurements from a dropped landmark $\ell_i$ only if the following conditions are satisfied: (i) the distance between $\ell_i$ and the vehicle is less than or equal to the sensing range, ${\rho}^{\max}$ and (ii) there exists a line of sight (LOS) between the vehicle and the landmark $\ell_i$. If either of the above conditions fail, then the vehicle does not obtain any range measurements from $\ell_i$. We let $\bm \rho$ represent the vector of range measurements obtained by the vehicle. Due to the existence of two known landmarks, $\ell_a$  and $\ell_b$ at the start of the tunnel, the vehicle starts receiving measurements from these two known landmarks as it starts its mission. As it traverses its path through the tunnel, landmarks are dropped and it receives additional range measurements. Similarly, as landmarks go out of range or lose LOS, they can no longer provide range measurements. Hence, the actual number measurements in $\bm \rho$ keeps varying dynamically depending on the landmarks that are within the vehicle's sensing range and have a LOS with the vehicle. For ease of exposition, we assume that $\bm \rho$ is a vector of fixed length of size $(n+2)$, where the first two elements denote the measurements from the known landmarks at the start of the tunnel. Given these notations, the measurement model is given by 
\begin{flalign}
    \bm \rho = 
    \begin{bmatrix} 
        \rho_a \\ \rho_b \\ \rho_1 \\ \vdots \\ \rho_n 
    \end{bmatrix} = 
    \begin{bmatrix}
        \|(x, y), (x_a, y_a)\|_2 \\
        \|(x, y), (x_b, y_b)\|_2 \\ 
        \|(x, y), (x_1, y_1)\|_2 \\
        \vdots \\
        \|(x, y), (x_n, y_n)\|_2
    \end{bmatrix} = \bm h(\bm x, \bm x_1, \dots, \bm x_n)
    \label{eq:measurement-model}
\end{flalign}
where, $\rho_a$ and $\rho_b$ are the range measurements from the two known landmarks, $\ell_a$ and $\ell_b$, at the start of the tunnel and $\rho_i$ for $i \in \{1, \dots, n\}$ denotes the range measurements obtained from landmark $\ell_i$. We remark that in Eq. \eqref{eq:measurement-model}, not all measurements are available at all instants of time and only a subset of measurements is at any time instant is available and they are alone used to perform the estimation.

\subsection{Extended Kalman Filter} \label{subsec:ekf}
Given the dynamics in Sec. \ref{subsec:models} and \ref{subsec:measurement}, we now present the state estimation algorithm. State estimation is required in this context due to the presence of noise and uncertainties in the dynamic model and measurement equations in Eq. \eqref{eq:uv-model} -- \eqref{eq:measurement-model}, respectively. For ease of exposition, we present the combined state space equations as follows: 
\begin{subequations}
\begin{flalign}
\dot{\bm X} &= \begin{bmatrix} \dot{\bm x} \\ \dot{\bm x}_1 \\ \vdots \\ \dot{\bm x}_n \end{bmatrix} = 
\begin{bmatrix}
    \bm f(\bm x, \bm u) + \bm s(t) \\ 0 \\ \vdots \\ 0
\end{bmatrix} \label{eq:state} \\
\bm Z_k &= \bm h(\bm X_k) + \bm w_k \label{eq:output}
\end{flalign}
\label{eq:state-space}
\end{subequations}
where, $\bm X$ is the vector of all the states of the vehicle and the landmark positions. The vehicle's states, $\bm x$, are the initial states in the estimator. The landmark drop positions $\bm x_i$ for every $i \in \{1, \dots, n\}$ are added dynamically to the vector $\bm X$ as and when they are deployed by the vehicle. The state equation in Eq. \eqref{eq:state} is in continuous time and the measurement equation in Eq. \eqref{eq:output} is in discrete time since is is typical for sensors to be sampled and processed in a digital hardware at a certain sample rate. In Eq. \eqref{eq:output}, $\bm X_k = \bm X(t_k)$. The terms $\bm s(t)$ and $\bm w_k$ are the process and measurement noise in the system. The process and measurement noise arise the uncertainties in the speed, yaw-rate of the vehicle and sensing noise, respectively. $\bm s(t)$ and $\bm w$ are assumed to be multivariate Gaussian random variables with covariances $\bm Q(t)$ and $\bm R_j$, respectively i.e., $\bm s(t) \sim \mathcal N(\bm 0, \bm Q(t))$ and $\bm w_k \sim \mathcal N(\bm 0, \bm R_k)$. Here, $\bm Q(t) = \operatorname{diag}(\sigma_v^2, \sigma_{\omega}^2)$ and $\bm R_k = \operatorname{diag}(\sigma_a^2, \sigma_b^2, \sigma_1^2, \dots, \sigma_n^2)$ where $\sigma_v$ and $\sigma_{\omega}$ are the standard deviations in the speed and the yaw-rate of the vehicle, and $\sigma_i$ is the standard deviation in the range measurement from landmark $\ell_i$. Furthermore, we also assume that $\sigma_a = \sigma_b = \sigma_1 = \dots = \sigma_n = \sigma_{\ell}$ this uncertainty in range measurement arises due to the on-board range sensor in the vehicle. Finally, we use $\hat{\bm X}_{q|m}$ and $\bm \Sigma_{q|m}$ to represent the estimate of $\hat{\bm X}$ and the covariance of the estimates at time, $q$, given observations up to and including at time $q \leqslant m$. 

We use an Extended Kalman Filter (EKF), a nonlinear version of the traditional Kalman filter \cite{beard2012small}, to compute $\hat{\bm X}$. Our approach differs from the traditional EKF in the sense that the estimates of the landmark drop locations are added dynamically to the initial state vector after it is dropped by the vehicle. Once the EKF is initialized, the main estimation procedure consists of two main steps (i) prediction step and (ii) measurement update step. In the prediction step, the state estimate and its associated covariance are predicted by propagating the dynamic models in Eq. \eqref{eq:uv-model} and \eqref{eq:landmark-model} and in the measurement step, these values are updated using the incoming measurements. In particular, we utilize a version of the EKF termed as the continuous-discrete EKF \cite{beard2012small}. In the continuous-discrete version of the EKF, measurement updates are performed at discrete time steps using a linearized version of the measurement equations. The prediction step propagates state estimate using the continuous dynamic models of the vehicle and the landmarks using finite difference; also, the covariance associated with the state estimates is propagated using a linearization of the continuous dynamic model. For the sake of completeness, the initialization, prediction, and measurement update equations are given below. 
\begin{subequations}
\begin{flalign}
    & \text{Initialization: } \notag \\
    & \hat{\bm X}_{0|0}=\mathbb{E}\bigl[\bm X(t_0)\bigr] \label{eq:intial_estimate} \\ 
    & \bm \Sigma_{0|0}=\mathbb{E}\left[\left(\bm X(t_0)-\hat{\bm X}(t_0)\right)\left(\bm X(t_0)-\hat{\bm X}(t_0)\right)^{\top}\right] \label{eq:initial_cv}
\end{flalign}
\label{eq:initialization}
\end{subequations}
As remarked previously, during initialization only the vehicle's states are contained in $\hat{\bm X}_{0|0}$ and $\bm \Sigma_{0|0}$. When landmarks are dropped, initial state estimates and the covariance for the landmark positions are appended at that time point dynamically to the state estimates and their covariance. As for the prediction step, at time $t_k$, the following differential equations are solved to obtain $\hat{\bm X}_{k|k-1}$ and $\hat{\bm \Sigma}_{k|k-1}$: 
\begin{subequations}
\begin{flalign}
    & \dot{\hat{\bm x}} = \bm f(\hat{\bm x}, \bm u),  \label{eq:uv-ode}  \\
    & \dot{\hat{\bm x}}_i = 0 \quad \forall \ell_i \text{ deployed, and} \label{eq:lm-ode} \\
    & \dot{\bm \Sigma} = \bm{F}\bm \Sigma+\bm \Sigma\bm F^{\top} + \bm {Q}. \label{eq:cv-ode}
\end{flalign}
\label{eq:prediction-odes}
\end{subequations}
In Eq. \eqref{eq:cv-ode}, $\bm F(t)$ is the Jacobian of the state equation. The initial conditions for the three differential equations are $\hat{\bm x}_{k-1|k-1}$, ${\hat{\bm x}_i}_{k-1|k-1}$, and $\bm \Sigma_{k-1|k-1}$, respectively. The differential equations are solved by using a finite-difference discretization scheme. Finally, the measurement update equations for a range measurement from landmark $\ell_i$, at time $t_k$ are given by 
\begin{subequations}
\begin{flalign}
    \bm K^i_k &= \bm \Sigma_{k|k-1} {\bm {H}^i_{k}}^{\top}\left(\bm H^i_{k}\bm \Sigma_{k|k-1}{\bm H^i_{k}}^{\top} + \bm R_{k}\right)^{-1}, \label{eq:kalman-gain} \\
    \hat{\bm X}_{k|k} &= \hat{\bm X}_{k|k-1} + \bm{K}^i_{k}\left(\bm{Z}_{k} - \bm h(\hat{\bm X}_{k|k-1})\right) \text{ and}\label{eq:state-update} \\
    \bm \Sigma_{k|k} &= (\bm I - \bm K^i_{k}\bm H^i_k)\bm \Sigma_{k|k-1}. \label{eq:cv-update}
\end{flalign}
\label{eq:measurement}
\end{subequations}
In Eq. \eqref{eq:measurement}, $\bm H_k^i$ is the measurement Jacobian corresponding to the measurement from landmark $\ell_i$ and $\bm K_k^i$ is the Kalman gain corresponding to that measurement. Next, we present a discussion on the observability of the dynamical system and consistency aspects of the EKF in relation to this problem setup. 

\subsection{Remark on observability of the system} \label{subsec:observability}
\review{It is known in the literature that using EKF to estimate states of an observable system works well in practice. But, in our case the state-space representation of the system presented in the previous section is unobservable if the path the vehicles traverses is not closed, i.e., the uncertainty corresponding to the state estimate will always be away from zero and will keep increasing. When the trajectory is closed, because we assume that at the start of the trajectory the vehicle has access to two known landmark positions, the system will be partially observable periodically on the path's loop-closure. The unobservability of the proposed system also follows from the unobservability results for SLAM \cite{lee2006observability}. Unobservability degrades the quality of the estimates provided by the EKF in the absence of external features to correct these estimates. The main idea of SLAM is to identify features in an environment, estimate the positions of the features and use these estimates to in turn localize the vehicle. But in feature-deficient environments like the ones considered in this article, we propose to inject features in the environment (landmarks). In the subsequent sections, we combine techniques from machine learning and mixed-integer optimization to inject landmarks at specific time intervals and this is combined with EKF to perform effective state estimation of this unobservable system. Similar to the setting of SLAM, the vehicle is localized using the position estimate of the injected features. Hence, the unobservability results extend in a straight-forward manner to the proposed approach as well.} Unobservability has direct implications on the uncertainty in position estimate of the EKF defined below. To formally define the position uncertainty, we let $\bm \Sigma_p(t)$ denote the vechicle-position covariance (sub-matrix of $\bm \Sigma(t)$ corresponding to the vehicle's state estimates $\hat x(t)$ and $\hat y(t)$) provided by the EKF, i.e., 
\begin{flalign}
 \bm \Sigma_p(t) = \begin{bmatrix} 
 \Sigma_{xx}(t) & \Sigma_{xy}(t) \\ 
 \Sigma_{xy}(t) & \Sigma_{yy}(t) 
\end{bmatrix}. \label{eq:position-covariance}
\end{flalign}
\review{Given the vehicle-position covariance matrix in Eq. \eqref{eq:position-covariance}, the instantaneous position error uncertainty in any given direction $\bm v$ is given by 
\begin{flalign}
    \sigma_{\bm v}(t) = \bm v^{\intercal} \cdot \bm \Sigma_p^{1/2}(t) \cdot \bm v \label{eq:v-error-uncertainty} 
\end{flalign}
where, $\bm \Sigma_p^{1/2}(t)$ is the square root of the position covariance matrix in Eq. \eqref{eq:position-covariance}. When $\bm v = [1 ~~ 0]^{\intercal}$ or $\bm v = [0 ~~ 1]^{\intercal}$, we get 
$$\sigma_x(t) = \left(\bm \Sigma_p^{1/2}(t)\right)_{xx} \text{ and } \sigma_y(t) = \left(\bm \Sigma_p^{1/2}(t)\right)_{yy}$$ 
where, $\sigma_x(t)$ and $\sigma_y(t)$ are the instantaneous position errors along the $x$ and $y$ directions, respectively.
Given this definition, the average instantaneous position uncertainty is computed using the following integral:
\begin{flalign}
    \int_0^{2\pi} \begin{bmatrix} \cos \theta \\ \sin \theta \end{bmatrix}^{\intercal} \cdot \bm \Sigma_p^{1/2}(t) \cdot \begin{bmatrix} \cos \theta \\ \sin \theta \end{bmatrix} ~d\theta = \frac 12 \cdot \operatorname{Trace} \bm \Sigma_p^{1/2}(t). \label{eq:integral-tr}
\end{flalign}
Given that the average instantaneous position uncertainty is $\frac 12 \cdot \operatorname{Trace} \bm \Sigma_p^{1/2}(t)$, we define the instantaneous position uncertainty, $P(t)$ as 
\begin{flalign}
    P(t) = \operatorname{Trace} \bm \Sigma_p^{1/2}(t). \label{eq:position-error}
\end{flalign}
Intuitively, $P(t)$ is proportional to the average instantaneous position error, where the average is taken over all possible directions. From the definition of $P(t)$ in Eq. \eqref{eq:position-error} combined with the fact that the state-space representation of the system in the previous section is unobservable, we can conclude that $P(t)$ will keep increasing as the vehicle navigates through the tunnel for a longer duration i.e., the longer the length of the tunnel the greater will be position uncertainty $P(t)$.  }
\begin{figure}
    \centering
    \includegraphics{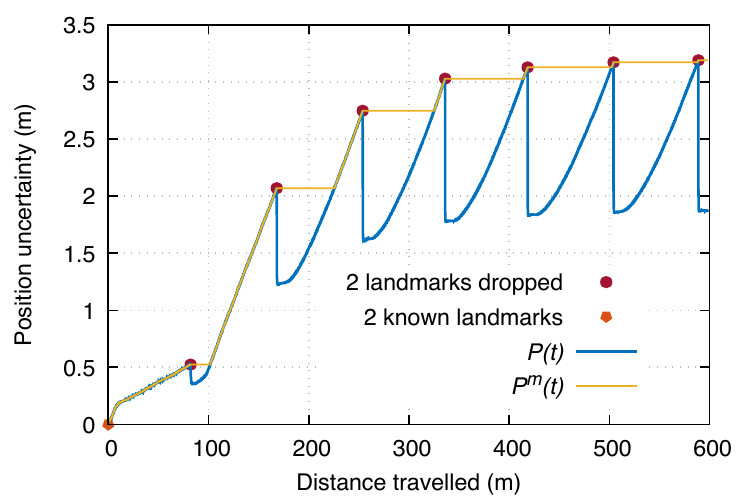}
    \caption{Instantaneous position uncertainty, $P(t)$ and $P^m(t)$, for the vehicle traveling along a tunnel of length $600$ \si{\meter} with speed $5$ \si{\meter\per\second}. The location two landmarks at the start of the tunnel is known without any uncertainty.}
    \label{fig:position-error-illustration}
\end{figure}
The Fig. \ref{fig:position-error-illustration} illustrates how $P(t)$ varies as the vehicle travels through a straight tunnel. $P(t)$ keeps increasing as the vehicle receives range measurement from two landmarks. As soon as additional two landmarks are dropped at the drop location, the position uncertainty decreases before it starts increasing again. We now define $P^m(t) = \max_{\tau \in [0, t]} P(\tau)$ as the maximum uncertainty in the position estimate upto time $t$. For an unobservable system the effective trend of $P(t)$ is to increase with $t$ i.e., $P^m(t)$ is not bounded from above. Nevertheless, deploying a larger number of landmarks or deploying landmarks more frequently will aid in decreasing the rate at which $P^m(t)$ increases as a function of $t$ as shown in Fig. \ref{fig:position-error-illustration}. The overall goal of the article is to compute the number of landmarks and their drop locations so that $P^{\max} = \max_{t\geqslant 0} P^m(t) \approx P_e$, the user-defined limit. In the next section, we qualitatively and quantitatively identify factors that impact $P^{\max}$ and develop a DNN to estimate the value of $P^{\max}$ for fixed values of these factors.

\section{Deep Neural Network (DNN) model} \label{sec:dnn}
In this section, we first try to identify factors that effect $P^{\max}$ when the vehicle is required to navigate through a straight tunnel with fixed length, $\mathcal L$. Throughout the rest of this section, we assume that the tunnel is straight with fixed length, $\mathcal L$.

\subsection{Factors affecting maximum position uncertainty} \label{subsec:features}
The model and measurement noise in the EKF (see Sec. \ref{subsec:ekf}) significantly affect the rate at which the instantaneous position uncertainty increases. In particular, speed of the vehicle ($v$), the standard deviation in the speed of the vehicle ($\sigma_v$), the sensing range ($\rho^{\max}$) of the vehicle's on-board range sensor, and the noise in the range measurement ($\sigma_{\ell}$) directly affect the $P(t)$ i.e., these values dictate the increase in $P^m(t)$ between successive landmark drops. Hence, these factors also indirectly influence the value of $P^{\max}$. 

Two other major factors that influence the increase in $P^m(t)$ are (i) the length of the tunnel and (ii) the frequency of landmark drops. It is clear from the Fig. \ref{fig:position-error-illustration} that as the length of the tunnel increases, the value of $P^m(t)$ keeps increasing. It is intuitively easy to see that the rate of increase of $P^m(t)$ will decrease with greater frequency of landmark drops. We mathematically characterize the frequency of landmark drops using the idea of drop distance between landmarks. Drop distance, $d$ is defined as 
\begin{flalign}
d = \rho^{\max} \cdot (1 - \lambda), \text{ where } \lambda \in [0, 1). \label{eq:drop-distance}
\end{flalign}
In Eq. \eqref{eq:drop-distance}, we refer to $\lambda$ as the overlap factor. The drop distance translates frequency of landmark drops to distance between two successive landmark drops. If $\lambda = 0$, the vehicle drops landmarks spaced exactly $\rho^{\max}$ distance apart i.e., the vehicle will only receive range measurements from exactly two landmarks during its path traversal. We also remark that if $\lambda > 1$, successive landmark drops are separated by a distance greater than $\rho^{\max}$; in this case, the vehicle would fly blind, i.e., without any range measurements, for a certain parts of the path traversal. For all other values of $\lambda \in (0, 1]$, the successive landmark drops are spaced within the sensing range i.e., the vehicle receives range measurements from at least four landmarks for travel distance of $\lambda \cdot \rho^{\max}$ after makes a landmark-drop. In summary, decreasing drop distance leads to better error and corresponding uncertainty estimates from the EKF and hence, lower $P^{\max}$. In summary, the six main mission parameters the affect the value of $P^{\max}$ are $(v, \sigma_v, \rho^{\max}, \sigma_{\ell}, \mathcal L, \lambda)$; in the subsequent section, we develop a DNN that can predict the value of $P^{\max}$ for a fixed value of mission parameters. 

\subsection{DNN design} \label{subsec:dnn-design}
\begin{figure}
    \centering
    \includegraphics[scale=0.2]{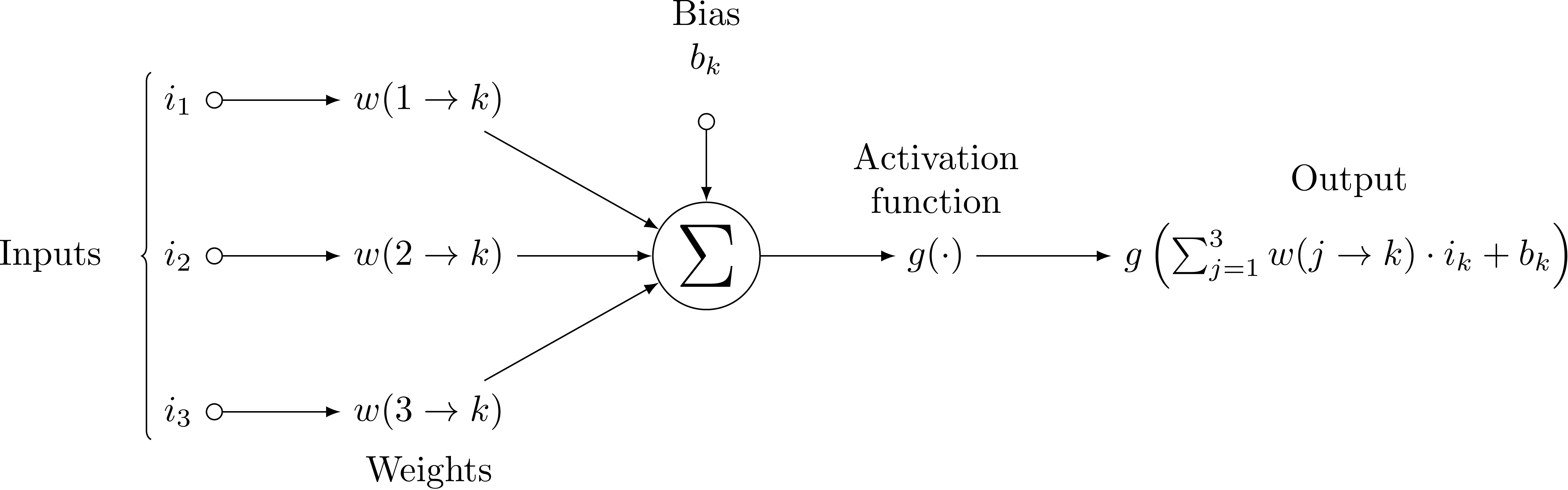}
    \caption{Input-output relationship for a single neuron $m_k$.}
    \label{fig:neuron}
\end{figure}

A DNN is an artificial neural network with multiple layers between the input and output layers \cite{bengio2009learning}. Each input and output in the DNN is represented using a node or a neuron in the input and output layer respectively. The input and output layers are in turn connected to each other through a finite number of hidden layers with each hidden layer consisting of a finite number of hidden nodes or neurons. Edges in the DNN go from a neuron in one layer to another neuron in the subsequent layer indicating the flow of information from one neuron to the other. Fundamentally, a neuron performs two operations (i) receives inputs from other neurons and combines them together and (ii) perform a mathematical operation on the combined value to obtain an output. In a DNN, operation (i) is the sum of the weighted linear combination of the inputs and a bias and the operation (ii) is referred to as an \textit{activation function}. Suppose that a neuron $m_k$ receives information $i_1$, $i_2$, $i_3$ from three neurons $m_1$, $m_2$, and $m_3$, respectively, then output of $m_k$ is as follows:
\begin{flalign}
g\left(\sum_{j = 1}^3 w(j \rightarrow k) \cdot i_k + b_k\right). \label{eq:output-neuron} 
\end{flalign}
In Eq. \eqref{eq:output-neuron}, $g(\cdot)$ is the activation function of the neuron, $b_k$ is referred to as the bias of the neuron and $w(j \rightarrow k)$ is referred to as the weight of the edge that connects neurons $n_j$ and $n_k$. \review{An illustrative figure of a single neuron is also shown in Fig. \ref{fig:neuron}.}
The nodes in the input layer do not perform any operation and only pass the information to the nodes or neurons in the next layer. In summary, DNN is simply made out of layers of neurons, connected in a way that the input of one layer of neuron is the output of the previous layer of neurons (after activation). Hence, the output of a DNN is completely determined by the weights of the edges connecting the nodes between successive layers and the biases of each hidden node. \review{We also remark that since we utilize DNNs only as a means to an end, we provide only a brief overview of the working principles of a DNN and we refer the readers to \cite{gurney2018introduction} for a more detailed presentation. }

For the purpose of this article, we use the Rectified Linear Unit ($\relu$) as the activation function $\relu(x) = \max \{0, x\}$. One main advantage of the $\relu$ activation function that we will put to effective use in the later sections is the fact that $\relu$ is a piecewise linear function. The input layer for our DNN consists of $6$ nodes, one for each $(v, \sigma_v, \rho^{\max}, \sigma_{\ell}, \mathcal L, \lambda)$. There is exactly one output neuron that represents $P^{\max}$. We also choose two hidden layers with each hidden layer consisting of $20$ neurons each. With the aforementioned DNN structure, we train the DNN; training entails computing the weights and biases of the edges and nodes based on a given input-output training data set. This involves defining a loss-function that evaluate a candidate solution. For our case, we utilize the Mean Squared Error (MSE) between the actual output and the output provided by the DNN and during training we aim to minimize the loss function. We train the DNN using a stochastic gradient descent algorithm to obtain the weights and biases that minimize the loss function. 

Normalization of the input and output data plays a very important role in extracting better performance from the DNN. In this work, we normalize both the input and the output data to lie in the range $[-1, 1]$ and the DNN is trained using the normalized data. During the testing phase, any general input value is always normalized and provided to the DNN and the output from the DNN is de-normalized to obtain the correct value. 

\subsection{Operation of a DNN as a mixed-integer linear program} \label{subsec:dnn-milp}
Once a DNN is trained, the operation of the DNN with $\relu$ activation functions can be equivalently formulated as a Mixed-Integer Linear Program (MILP) \cite{fischetti2017deep}. This in turn provides an efficient way to solve inverse problems using the DNN. In particular, we seek to answer the question: \\

\noindent \textit{Given values of $(v, \sigma_v, \rho^{\max}, \sigma_{\ell}, \mathcal L)$, what is the value of $\lambda$ (overlap factor) that can minimize $(P^{\max} - P_e)^2$?} \\

\noindent The objective $(P^{\max} - P_e)^2$ seeks to keep $P^{\max}$ as close as possible to the user-defined value of $P_e$. Before we present the reformulation of the DNN into an MILP, we note that all the operations that occur within a DNN can be expressed using linear equations but for the $\relu$ activation function. 
%
%
The $\relu$ activation function is a piecewise linear function. Authors in \cite{grimstad2019relu} show that this piecewise linear function can be reformulated into a set of linear constraints by the introduction of a binary variable. Hence, the DNN with $\relu$ activation function can be equivalently converted into a MILP with the number of binary variables being equal to the number of nodes with a $\relu$ activation in the DNN. For the sake of completeness, we present the MILP formulation and in order to present the MILP formulation, we present some additional notations. We present the formulation for the DNN considered in this article with two hidden layers, one input layer and one output layer. In total the DNN has $4$ layers numbered $0$ - $3$ with $n_0 = 6$, $n_1 = 20$, $n_2 = 20$, and $n_3 = 1$ nodes in each layer, respectively. The $6$ nodes in the input layer (layer $0$) denote the $6$ inputs to the DNN given by $(v, \sigma_v, \rho^{\max}, \sigma_{\ell}, \mathcal L, \lambda)$. The node in the output layer (layer $3$) represents $P^{\max}$. The nodes in all layers, but for the input layer, are assumed to be neurons with a $\relu$ activation function. We let $N(i, j)$ the node $i$ in the layer $j$. Each layer $k \in \{1, 2, 3\}$ is associated with weight and bias vector $\bm W^k \in \mathbb R^{n_k \times n_{k-1}}$ and $\bm b^k \in \mathbb R^{n_k}$ (the bias value for the output node is assumed to be $0$). The action of the DNN represents a function whose domain and range are $\mathbb R^{n_0}$ and $\mathbb R$, respectively. 

For each layer $k \in \{0, 1, 2, 3\}$, we let $\bm y^k \in \mathbb R^{n_k}$ and $y^k_j$ denote the vector of outputs from that layer and the output from the node $N(k, j)$, respectively. Then for each layer in $k \in \{1, 2, 3\}$, the following relationship holds: 
\begin{flalign}
    \bm y^k = \relu\left((\bm W^k)^{\top} \bm y^{k-1} + \bm b^k\right). \label{eq:dnn-eq}
\end{flalign}
In the above equation, the $\relu$ function is applied to the vector of outputs componentwise. Eq. \eqref{eq:dnn-eq} can be converted to a linear constraint system by the introduction of two sets of variables for every for $k \in \{1, 2, 3\}$: (i) $\bm y^{k}_- \in \mathbb R^{n_k}$  and (ii) $\bm z^k \in \{0, 1\}^{n_k}$. Using these two sets of variables, Eq. \eqref{eq:dnn-eq} can be equivalently represented using the following constraint system for each layer in $k \in \{1, 2, 3\}$: 
\begin{subequations}
\begin{flalign}
    \left((\bm W^k)^{\top} \bm y^{k-1} + \bm b^k\right) = \bm y^k - \bm y^k_-,  \label{eq:dnn-relu-1} \\
    \bm y^k \leqslant M \bm z^k, ~ \bm y^k_- \leqslant -M (\bm 1 - \bm z^k),  \label{eq:dnn-relu-2} \\ 
    \bm y^k \geqslant 0, ~ \bm y^k_- \geqslant 0, ~ \bm z^k \in \{0, 1\}^{n_k}. \label{eq:dnn-relu-3}
\end{flalign}
\label{eq:relu-reformulation}
\end{subequations}
In Eq. \eqref{eq:dnn-relu-2}, $M$ is a large positive constant and $\bm 1$ is a vector of ones of appropriate dimension. It is trivial to see the equivalence between Eq. \eqref{eq:dnn-eq} and \eqref{eq:relu-reformulation} by examining the consequences of the value of $(\bm W^k)^{\top} \bm y^{k-1} + \bm b^k$ in both the equation systems. An interested reader is refered to \cite{fischetti2017deep,grimstad2019relu} for detailed expositions on their equivalence. 

We now formulate the inverse problem of finding the value of $\lambda$ that can minimize $(P^{\max} - P_e)^2$ constrained by the DNN. The problem is formulated as a MILP with a convex quadratic objective function as follows:
\begin{subequations}
\begin{flalign}
    \min: ~~ & (P^{\max} - P_e)^2 \text{ subject to: Eq. \eqref{eq:relu-reformulation},}  \label{eq:inverse-obj} \\
    & \begin{bmatrix} v \\ \sigma_v \\ \rho^{\max} \\ \sigma_{\ell} \\ \mathcal L \end{bmatrix} = 
    \begin{bmatrix} v^* \\ \sigma_v^* \\ \rho^{\max*} \\ \sigma_{\ell}^* \\ \mathcal L^* \end{bmatrix} = 
    \begin{bmatrix} y^0_1 \\ y^0_2 \\ y^0_3 \\ y^0_4 \\ y^0_5 \end{bmatrix},
    \label{eq:inverse-inputs} \\ 
    & y^0_6 = \lambda, ~~ 0 \leqslant \lambda \leqslant 1 - \varepsilon, ~~ y^3_1 = P^{\max}. \label{eq:inverse-outputs} 
\end{flalign}
\label{eq:inverse}
\end{subequations}
The constraint in Eq. \eqref{eq:inverse-inputs} fixes the values for the variables $(v, \sigma_v, \rho^{\max}, \sigma_{\ell}, \mathcal L)$ based on mission parameters and links then to the first $5$ inputs to the DNN. Eq. \eqref{eq:inverse-outputs} assigns the value of $\lambda \in [0, 1)$ to be the last input and $P^{\max}$ to the value of the output of the DNN. The DNN's MILP reformulation is also added to the problem in Eq. \eqref{eq:inverse-obj}. The optimization problem in Eq. \eqref{eq:inverse} can be readily solved by off-the-shelf commercial and open-source MILP solvers within a fraction of a second. It results in a optimal value of $\lambda^*$ and an estimate of the maximum position uncertainty $P^{\max*}$. In the next section, we present algorithms to adjust the drop locations obtained by solving the MILP in Eq. \eqref{eq:inverse}. 

\section{Landmark drop adjustment algorithms} \label{sec:drop-adjustment}
The MILP presented in the previous section results in a value of $\lambda^*$ so that the value or $P^{\max*} \approx P_e$ when the tunnel is a straight with length $\mathcal L^*$ and the other vehicle and sensor parameters are $v^*$, $\sigma_v^*$, and  $\rho^{\max*}$. This value of $\lambda^*$ can be converted to landmark drop locations by computing the corresponding drop distance $d^* = \rho^{\max*} (1 - \lambda^*)$ i.e., two landmarks on the either side of the vehicle's path is placed at every $d^*$ units from the start of the tunnel. When the tunnel is not straight and has turns, the drop distance, $d^*$, provided by the MILP may not be sufficient to ensure that $P^{\max*} \approx P_e$. This section seeks to address this issue by developing landmark drop adjustment algorithms that adjust the drop locations provided by the MILP suitably and adds more landmark drops to ensure $P^{\max*} \approx P_e$.

\subsection{Straight tunnel with known length} \label{subsec:straight}
We recall that measurement model in Sec. \ref{subsec:measurement} assumes that the vehicle can obtain range measurements from a dropped landmark if and only if the landmark lies within the sensing range of the vehicle and there exists a LOS between the vehicle and the landmark. In a straight tunnel, there is no LOS loss and hence, the drop distance provided by the MILP is sufficient to keep the $P^{\max} \approx P_e$ throughout the path traversal. Hence, in this case the number of landmarks that need to be carried by the UV is given by 
\begin{flalign}
{\mathcal N}_{\text{straight}} = 2 \cdot \left\lfloor \frac{\mathcal L^*}{d^*} \right\rfloor. \label{eq:straight}
\end{flalign}
The above landmark count does not include the two known landmarks that are at the start of the tunnel and assumes two landmarks are dropped at every drop location, one on either side of the vehicle's path. 

\subsection{Tunnel with known length and number of turns} \label{subsec:known-turns}
\begin{figure}
    \centering
    \includegraphics{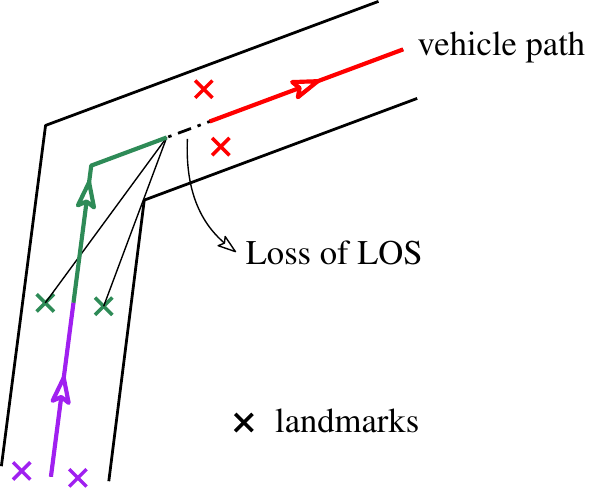}
    \caption{Loss of LOS along the vehicle path during a turn. The vehicle loses LOS from the green landmarks even though they are within the sensing range of the vehicle due to the turn. The vehicle's direction of travel is shown using the arrows.}
    \label{fig:los-loss}
\end{figure}

In this section, we assume that we are given a tunnel with turns and apart from knowing $\mathcal L^*$, we additionally know the number of turns $\mathcal T$ in the tunnel. In this case, we first assume that the landmarks will be placed at every $d^* = \rho^{\max*} (1 - \lambda^*)$ ($\lambda^*$ is obtained by solving the MILP with tunnel length $\mathcal L^*$) units along the vehicle's path. A turn in the tunnel can cause a loss of line of sight (LOS) between the range sensor in the vehicle the landmark as soon as the the vehicle makes the turn (see Fig. \ref{fig:los-loss} for an illustration). This can lead to the $P^{max}$ taking a much higher value than desired by the user.  
This loss in LOS can be removed if two additional landmarks are dropped as the the vehicle negotiates. But this will increase the number of landmarks that the vehicle has to carry. Nevertheless, given this level of information on the tunnel and the requirement that the vehicle always needs range measurements from two landmarks to keep $P^{\max}$ in check, this provides a conservative upper bound on the number of landmarks that the vehicle has to carry. As we shall see later in the results, the greater the number of turns and the shorter the distance between the turns, the $P^{\max}$ will take a value much lower than $P_e$ because of the conservativeness of this approach. In summary, the number of landmarks is given by 
\begin{flalign}
{\mathcal N}_{\text{turns}} = {\mathcal N}_{\text{straight}} + 2 \cdot \mathcal T. \label{eq:turns}
\end{flalign}
In the next section, we see that if we know the full topology of the tunnel, then the number of landmarks can be reduced to a value close to ${\mathcal N}_{\text{straight}}$. 

\subsection{Tunnel with known topology}  \label{subsec:known-topology} Here, we assume that full topology of the tunnel is known i.e., the length of the tunnel, the number of turns, the angle of each turn, and the length of the vehicle's path between successive turns. Given this information, we develop an algorithm to determine the number of landmarks that the vehicle has to carry and their drop locations to keep $P^{\max}$ close to $P_e$. To that end, we start by with a potential drop location that drops two landmarks every $d^*$ units along the length of the vehicle's path. Let $\mathcal T = \{\tau_1, \dots, \tau_m\}$ denote the set of turns. For each turn $\tau_i$, as mentioned in the previous section, the vehicle can lose LOS from the last two landmarks that were dropped before the turn. The pseudo-code of the algorithm to compute the updated set of drop locations is given in Algorithm \ref{algo:adjustment}.
\begin{algorithm}
\caption{Drop location adjustment algorithm}
\label{algo:adjustment}
\hspace*{\algorithmicindent} \textbf{Input} - $\mathcal T = \{\tau_1, \dots, \tau_m\}$, $d^*$, $\mathcal L^*$, vehicle path $\mathcal P$  \\
\hspace*{\algorithmicindent} \textbf{Output} - updated drop locations along $\mathcal P$
\begin{algorithmic}[1]
\For{turn $\tau$ in $\mathcal T$} 
    \State $\ell(\tau, \mathcal P) = (\ell^{\text{left}}, \ell^{\text{right}})$ \Comment{landmark pair before $\tau$} \label{step:landmark-pair}
    \State $\ell^{\text{left}} \gets $ last landmark to the left of $\mathcal P$ before $\tau$ \label{step:left}
    \State $\ell^{\text{right}} \gets $ last landmark to the right of $\mathcal P$ before $\tau$ \label{step:right}
    \State $d^p \gets ~$\textsc{ComputePullDistance}($\ell^{\text{left}}$, $\ell^{\text{right}}$, $\mathcal P$, $\tau$) \label{step:pull-distance}
    \State pull drop distances of landmarks after $\ell(\tau, \mathcal P)$ by $d^p$ \label{step:decrese}   \State add landmark drops at the end (if required) \label{step:adjust}
\EndFor
\end{algorithmic}
\end{algorithm}

\begin{figure}
    \centering
    \includegraphics[scale=0.8]{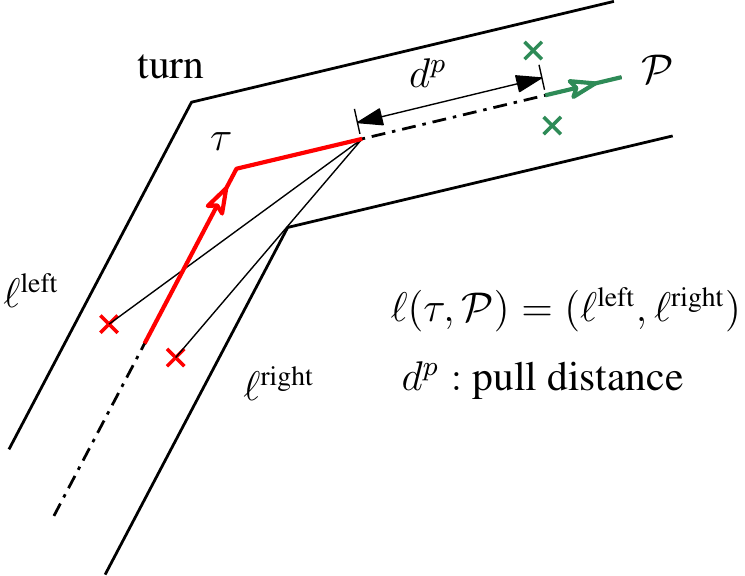}
    \caption{Illustration of the steps involved in Algorithm \ref{algo:adjustment}. The vehicle receives range measurements from other landmarks (not shown) as it traverses the path before the red landmarks are dropped.}
    \label{fig:pull}
\end{figure}

\begin{figure}
    \centering
    \includegraphics[scale=0.8]{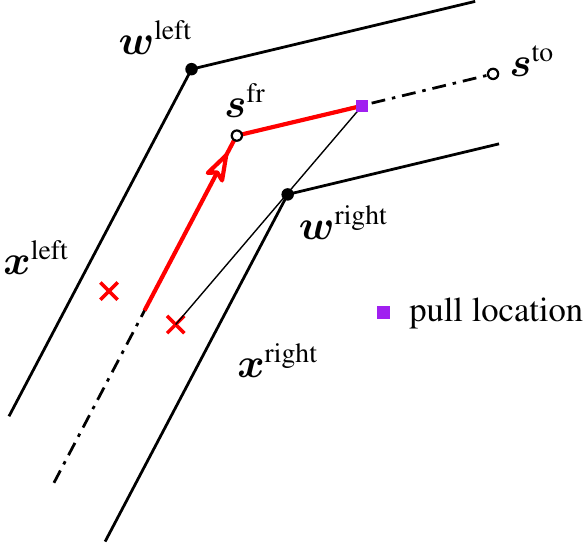}
    \caption{Notations for the algorithm to compute pull location. Here, the turn is to the right and hence, LOS will be lost from the right landmark first before the left landmark.}
    \label{fig:pull-location}
\end{figure}

In Step \ref{step:landmark-pair} of the Algorithm \ref{algo:adjustment}, $\ell(\tau, \mathcal P)$ denotes the landmarks along the vehicle's path that lie before the turn $\tau$. This is a pair of landmarks one to the left of the path, $\mathcal P$, and another to the right of $\mathcal P$ (see Fig. \ref{fig:pull}). In Step \ref{step:pull-distance}, the pull distance, $d^p$ is computed for the turn $\tau$ using the landmark drop locations of $\ell^{\text{left}}$ and $\ell^{\text{right}}$. This operation is encapsulated into the function \textsc{ComputePullDistance}. To formally define pull distance, we first define a pull location of turn $\tau$. The pull location of turn $\tau$ is the location along the path of the vehicle, $\mathcal P$, after the turn $\tau$ where the LOS between $\ell^{\text{left}}$ or $\ell^{\text{right}}$ and the range sensor in the vehicle is first lost. Once we compute the pull location, the pull distance is simply defined at the distance between the pull location and the next landmark drop location (drop location after $\ell(\tau, \mathcal P)$) along $\mathcal P$ (the green landmarks in the Fig. \ref{fig:pull}). The pull distance and the pull location are illustrated in Fig. \ref{fig:pull}. After the computation of the pull distance, all the landmarks drop locations after $\ell(\tau, \mathcal P)$ are pulled back by $d^p$ units (Step \ref{step:decrese}). During this process, a decrease in the overlap in the end of the vehicle's path may occur i.e., the drop distances between successive landmark drops at the end of the vehicle's path can become greater than $d^*$. This will lead to an increase in $P^{\max}$. To address this issue, we keep track of drop distances between successive landmarks after every pull back and if this drop distance is strictly greater then $d^*$, we add additional landmark drops to decrease it to exactly $d^*$ (Step \ref{step:adjust}). If we let this number of additional landmark drop locations be denoted by $\mathcal N_{\text{additional}}$, then the total number of landmark drops is given by 
\begin{flalign}
\mathcal N_{\text{full}} = \mathcal N_{\text{straight}} + 2 \cdot \mathcal N_{\text{additional}}. \label{eq:full} 
\end{flalign}
We now present the algorithm to compute the pull location, thereby completing the presentation of the Algorithm \ref{algo:adjustment}. To that end, it is easy to see that the LOS loss due to a turn to the left (right) will first result from the landmark to the left (right), $\ell^{\text{left}}$ ($\ell^{\text{right}}$). We present the algorithm for one turn and it remains the same for the rest of the turns. The problem of finding the point where that LOS is can be formulated as the solution to a linear system of equations. We first introduce some notations that will aid in formulating the problem. We let $\bm x^{\text{left}}$ and $\bm x^{\text{right}}$ denote the positions of the landmarks $\ell^{\text{left}}$ and $\ell^{\text{right}}$, respectively. We also let $\bm w^{\text{left}}$ and $\bm w^{\text{right}}$ denote the coordinates of the wall where the turn $\tau$ occurs. Finally, we let $\bm s^{\text{fr}}$ and $\bm s^{\text{to}}$ denote the start and end of the straight path after turn $\tau$. The Fig. \ref{fig:pull-location} illustrates all the notation introduced thus far. Given these notation, the pull location for a turn to the right (the system of equation is similar for a turn to the left) is computed as a solution to the following linear system 
\begin{subequations}
\begin{flalign}
& \alpha^{\text{right}} \geqslant 0, ~~ 0 \leqslant \beta^{\text{right}} \leqslant 1, \label{eq:multiplier-bounds} \\
& \bm x^{\text{right}} + \alpha^{\text{right}} \frac{\bm w^{\text{right}}- \bm x^{\text{right}}}{\|\bm w^{\text{right}}- \bm x^{\text{right}}\|} = \beta^{\text{right}} \bm s^{\text{fr}} + (1 - \beta^{\text{right}}) \bm s^{\text{to}}. \label{eq:ray-segment}
\end{flalign} 
\label{eq:right-pull}
\end{subequations}
The LHS of Eq. \eqref{eq:ray-segment} is the equation of the ray originating from $\bm x^{\text{right}}$ and directed towards $\bm w^{\text{right}}$ and the RHS gives the equation of the line segment joining $\bm s^{\text{fr}}$ and $\bm s^{\text{to}}$. If a solution exists for the above system of equations, then the solution gives the required pull location. The system of equations if the turn is to the left is similar and hence, it is not presented. Using the pull location, the pull distance, $d^p$, is computed as the distance between the pull location and the next landmark drop location along the path (see Fig. \ref{fig:pull}). We remark that if the next landmark drop location occurs before the pull location, then no pulling of the landmarks is performed. 

In summary based on the information that is available a-priori on the tunnel configuration, we have presented algorithms to optimize the number of landmarks and their drop locations. The least number of landmark drops is required when the tunnel is a straight with a pre-specified length i.e., $\mathcal N_{\text{straight}}$. When the full topology of the tunnel is known, the algorithm in Sec. \ref{subsec:known-topology} requires $\mathcal N_{\text{full}}$ number of landmarks and when only the number of turns with the length of the tunnel is known, the algorithm in Sec. \ref{subsec:known-turns} requires $\mathcal N_{\text{turns}}$ number of landmarks. \review{In summary, the results provided by the three algorithms are related to each as other as
\begin{flalign}
\mathcal N_{\text{straight}} \leqslant \mathcal N_{\text{full}} \leqslant \mathcal N_{\text{turns}} \label{eq:turn-relation}
\end{flalign}}
In the next section, we detail some failure points of the algorithms presented thus far and introduce novel methodologies to tackle these failure points by the use of additional sensors.

\section{Other sensors and system architecture} \label{sec:arch} 
The state estimation, the MILP and the landmark drop adjustment algorithms, presented thus far, can be glued together to form an efficient navigation system for the vehicle through feature-deficient environments like tunnels or mines. But the system built just using these approaches will not always guarantee that during the vehicle's entire path traversal the position uncertainty remains small or $P^{\max} \approx P_e$, primarily due to the following reasons (i) bias in the state estimation algorithm and (ii) range measurements are available effectively only from one landmark. This section presents a discussion of these two points of failure for the navigation system and develops methodologies to address the same. 


Bias in the state estimation algorithm is caused due to a drift in the state estimates provided by the algorithm. When the vehicle is traveling between way points, the drift is caused in the direction perpendicular to the direction of travel. This can cause the position error and the associated uncertainty to grow with time eventually leading to inconsistency of the EKF. Intuitively, if bias exists and proper bias modeling and handling mechanisms are not used for estimation to actively take bias into account, it can make a filter overconfident i.e., the estimation errors may grow beyond the associated uncertainty bounds with time which can make the filter inconsistent. Inconsistency of an EKF is caused when linearization of the system model is performed at incorrect state estimates. Primarily, the reason for the drift, and hence the bias, is that the heading control of the vehicle that has access to range measurements is indirect i.e., the errors in position estimates of the vehicle indirectly control the error in heading. To address this issue for unmanned vehicles, several techniques have been proposed in the literature ranging from bias-estimation, bias-correction \cite{simon2006optimal}, equipping the vehicle with bearing or heading sensor \cite{sharma2012graph} to name a few. In this article, we resort to a simpler approach of using two sensors on the vehicle that can measure the distance of the vehicle from the wall on the either side of tunnel. From here on, we refer to these sensors as \textit{wall-update} sensors. 

\begin{figure}
    \centering
    \includegraphics[scale=0.8]{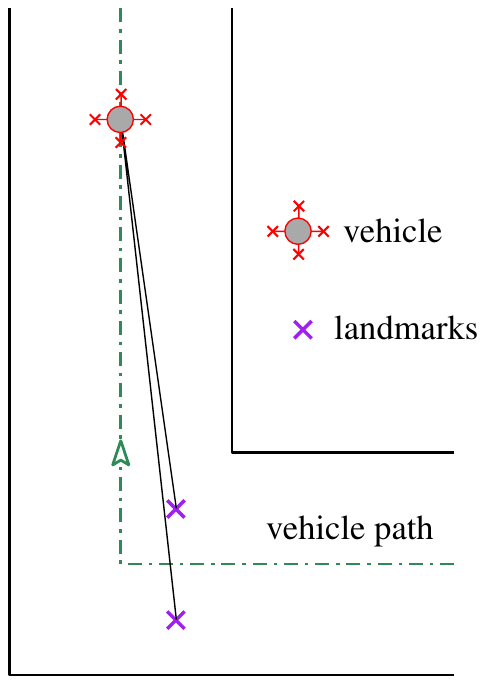}
    \caption{Illustration of a case where the two landmarks and the vehicle are effectively co-linear due to the noise in measurements.}
    \label{fig:l-obs}
\end{figure}

The second point of failure for the navigation system arises from the fact that though we make sure range measurements from at least two landmarks are available to the vehicle during its full path traversal, there can be situations the landmarks and the vehicle can become effectively co-linear due to large noise in measurements or low quality sensing equipment, resulting in a faster increase of position uncertainties. The Fig. \ref{fig:l-obs} shows a case where the two landmarks that provide the vehicle with range measurements are almost co-linear. This is not an issue if the measurement noise is small and if the landmark locations known, but in cases where there is huge measurement noise and the landmark locations are themselves estimated with uncertainties, this can create a large uncertainty in the position of the vehicle in the direction perpendicular to the vehicle's path. The wall-update sensors proposed in the next few paragraphs aid in addressing this issue as well. 

\subsection{Wall-update sensors} \label{subsec:wall-update}
The use of wall-update sensors is novel in the context of developing range-based navigation algorithms in feature-deficient algorithms. It simultaneously tackles the two issues elucidated in the previous paragraphs. The wall-update sensors measure the distance between the vehicle and the walls on the either side of the tunnel. These measurements are utilized to correct the heading through an inner control loop that gets updated at a lower frequency than update frequency of the state estimator. These measurements are utilized to ensure that the vehicle is always traversing along its path equidistant from the two walls on its either side. This in turn ensures that the heading errors along the vehicle's path traversal are close the zero, eliminates the drift in the vehicle's position estimate and thereby makes the EKF presented in Sec. \ref{sec:state-estimation} unbiased. In the subsequent sections, we show the importance of these wall-update sensors through extensive simulation experiments.

\subsection{Navigation system workflow} \label{sec:workflow}
\begin{figure}
    \centering
    \includegraphics[scale=0.8]{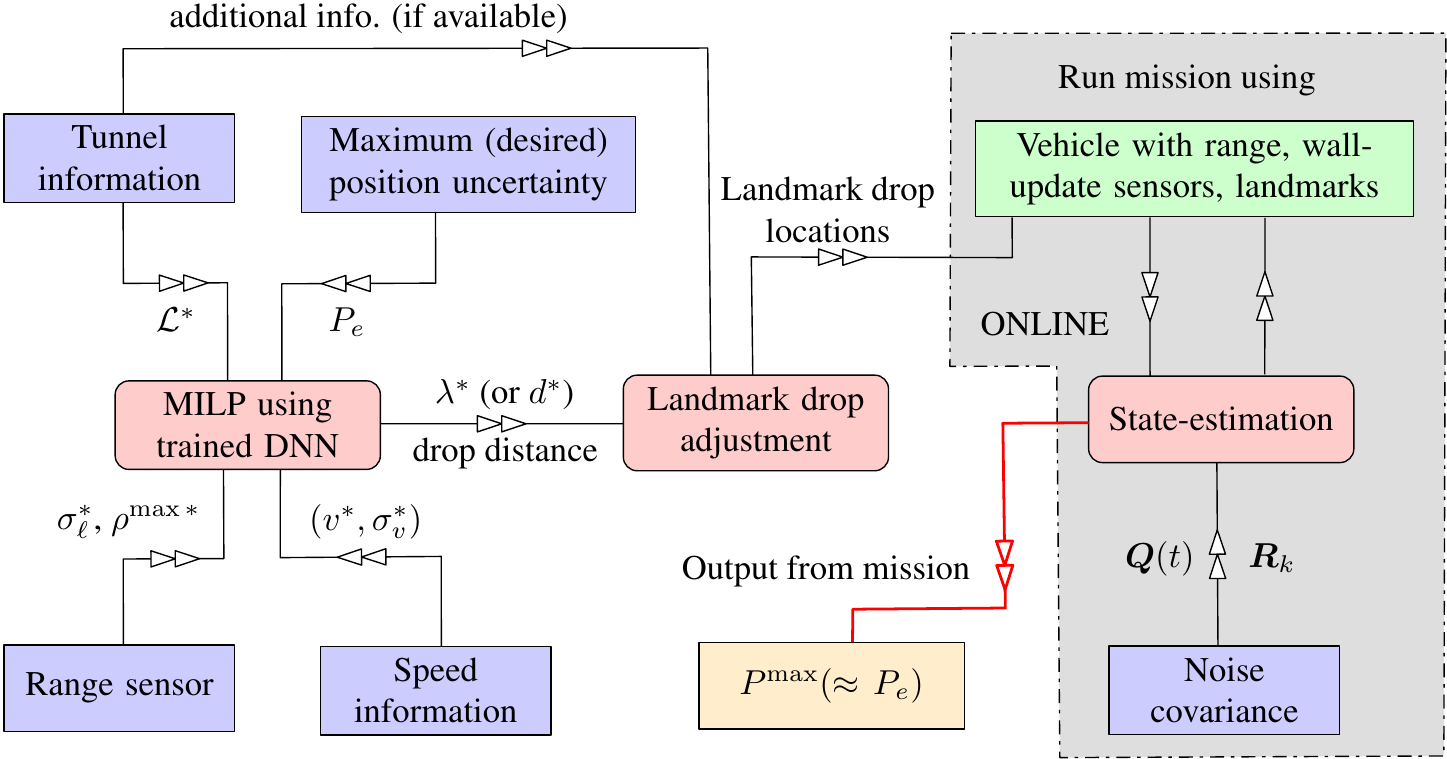}
    \caption{Workflow for the proposed navigation system. Red boxes denote the algorithms and the blue boxes are the inputs to the algorithms.}
    \label{fig:workflow}
\end{figure}

We now present an overview of the overall workflow of the navigation system. This would aid a user of this navigation system in understanding how the different technical aspects of this article are glued together. The flowchart in Fig. \ref{fig:workflow} provides an overview of the workflow. 

In the next section, we present simulation results that corroborate the effectiveness of all the algorithms presented in this article and of the overall navigation system for different sets of input parameters and tunnel configurations.

\section{Simulation results} \label{sec:results}
In this section, we present extensive simulation results that corroborate the effectiveness of the overall navigation system and each part separately. As far as implementation is concerned, the state estimation algorithm and the DNN model were built using MATLAB \cite{MATLAB} and the drop adjustment algorithms were implemented using the Julia programming language \cite{bezanson2012julia}. All simulation experiments were run on a Microsoft Windows computer equipped with Intel i7 processor, 16 GB RAM, and Nvidia GTX 1080 Ti graphic card. We begin by demonstrating the performance of the state estimation algorithms. 

\subsection{Performance of the state-estimation algorithm} \label{subsec:ekf-performance} 
\begin{figure}[!ht]
    \centering
    \includegraphics{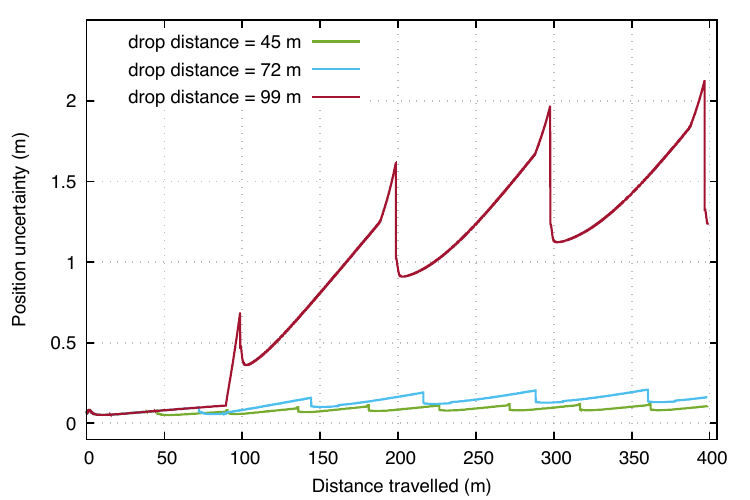}
    \caption{EKF performance for the vehicle traveling through a straight tunnel of length $400$ \si{\meter}. Observe that the rate of increase instantaneous position uncertainty increases with the landmark drop distance.}
    \label{fig:ekf-performance}
\end{figure}

For this set of results, we assume that the tunnel is straight with length $\mathcal L = 400$ \si{\meter}. We also assume the vehicle travels with a speed $v = 3$ \si{\meter\per\second} and a standard deviation of $\sigma_v = 0.3$ \si{\meter\per\second}; the other parameters are given by $\rho^{\max} = 90$ \si{\meter} and $\sigma_{\ell} = 0.1$ \si{\meter}. Two landmarks with known positions are available at the start of the tunnel symmetrically on the either side of the vehicle's path. As the vehicle navigates through the tunnel, we assume that two landmarks at a distance of $10$ \si{\meter} on the either side of the vehicle, respectively. The vehicle drops two landmarks periodically with a drop distance of $d$ \si{\meter}. The Fig. \ref{fig:ekf-performance} illustrates the instantaneous position uncertainty as the navigates through the straight tunnel for values of $d$ in the set $\{45, 72, 99\}$. \review{Here, the instantaneous position uncertainty is computed using Eq. \eqref{eq:position-error}.}  When $d = 99$ \si{\meter}, the vehicle will fly blind in certain parts of the tunnel, i.e., periodically it will not receive any range measurements for any dropped landmarks and hence, the position uncertainty will increase at a fast rate (see Fig. \ref{fig:ekf-performance}). For the other two values of $d$ i.e., $72$ \si{\meter} and $45$ \si{\meter}, the instantaneous position uncertainty is fairly small since, for certain parts of the vehicle's path, it receives range measurements from at least $4$ landmarks thereby enabling the EKF to keep the position uncertainty in check. 

\subsection{Performance of the DNN} \label{subsec:dnn-performance} 
\begin{figure}[ht!]
    \centering
    \includegraphics{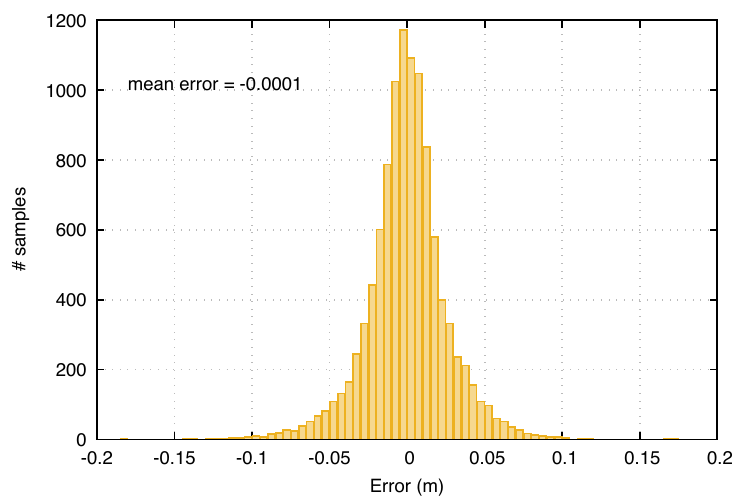}
    \caption{Histogram of errors produced by the DNN on the test data set.}
    \label{fig:dnn-performance}
\end{figure}

As remarked in Sec. \ref{subsec:dnn-milp}, the DNN proposed in this article consists of $6$ inputs in the input layer, one output in the output layer and two hidden layers with $20$ nodes in each hidden layers. Each node other than the ones in the input layer are equipped with the $\relu$ activation function. The training, validation, and testing data sets are randomly generated. In total, we generate $110000$ input-output samples with a $(70, 20, 10)\%$ training, validation, and testing split; the actual samples that go into this split are randomly chosen. To generate each sample, we first randomly choose input values from the sets shown in Table \ref{tab:inputs}. For all the cases, we assume that the tunnel is straight without any turns. To obtain the output corresponding to an input, we run the state estimation algorithm and compute the $P^{\max}$ provided by the simulation. For this entire process, MATLAB's Deep Learning Toolbox (\url{https://www.mathworks.com/help/deeplearning/index.html}) is used to train, analyze, and evaluate the DNN's performance.
\begin{table}
    \caption{Data for the DNN is generated by choosing a random value in the domain of the input.}
    \label{tab:inputs}
    \centering
    \begin{tabular}{c|c|c}
        \toprule 
        input & domain & units \\ 
        \midrule
        $v$ & $\{2, 2.5, 3, \cdots, 5\}$ & \si{\meter\per\second}\\
        $\sigma_v$ & $\{0.1 \cdot v, 0.2 \cdot v\}$ & \si{\meter\per\second}\\
        $\rho^{\max}$ & $\{50, 55, 60, \cdots, 100\}$ & \si{\meter}\\ 
        $\sigma_{\ell}$ & $\{0.1, 0.6, 1.1, 1.6, 2.1\}$ & \si{\meter}\\ 
        $\lambda$ &  $\{0, 0.05, 0.1, \cdots, 0.7\}$ & --\\
        $\mathcal L$ & $[0.2, 600]$ & \si{\meter}\\
        \bottomrule
    \end{tabular}
\end{table}
The DNN is trained for $10000$ epochs using Stochastic Conjugate Gradient (SCG) running on the graphic card. The error performance of the DNN on the test data set is show in the Fig. \ref{fig:dnn-performance}. 
\subsection{Effectiveness of MILP} \label{subsec:milp-performance}
\begin{figure}[ht!]
    \centering
    \includegraphics{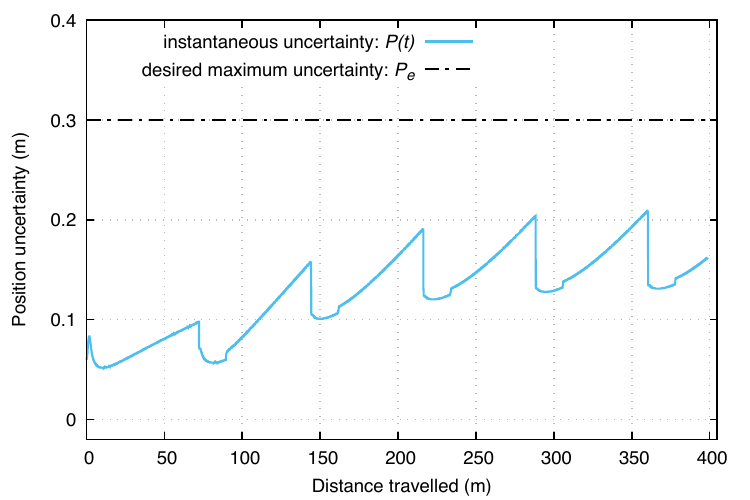}
    \caption{Instantaneous position uncertainty as the vehicle navigates through a straight tunnel using the drop distance computed by the MILP.}
    \label{fig:straight-performance}
\end{figure}

\review{In this section, we corroborate the effectiveness of the solution obtained by solving the MILP in keeping the maximum position uncertainty within its user-defined limit $P_e$. For this set of results, we consider a straight tunnel with length $\mathcal L^* = 400$ \si{\meter}. Other parameters for this experiment are as follows: $v^* = 3$ \si{\meter\per\second}, $\sigma_v^* = 0.3$ \si{\meter\per\second}, $\rho^{\max*} = 90$ \si{\meter} and $\sigma_{\ell}^* = 0.1$ \si{\meter}. Furthermore, the value of $P_e$, the desired maximum position uncertainty, is set to $0.3$. The problem is solved using CPLEX as the MILP solver in less than a second. The MILP results in a drop distance $d^* = 72$ \si{\meter}. For this value of drop distance, the instantaneous position uncertainty is shown in Fig. \ref{fig:straight-performance}. As seen from the Fig. \ref{fig:straight-performance}, the maximum position uncertainty for the full vehicle's path traversal lies well within its user-defined limit of $0.3$. }

\subsection{Effectiveness of landmark drop adjustment algorithms} \label{subsec:drop-adjustment-performance} 
\begin{figure}
    \centering
    \includegraphics{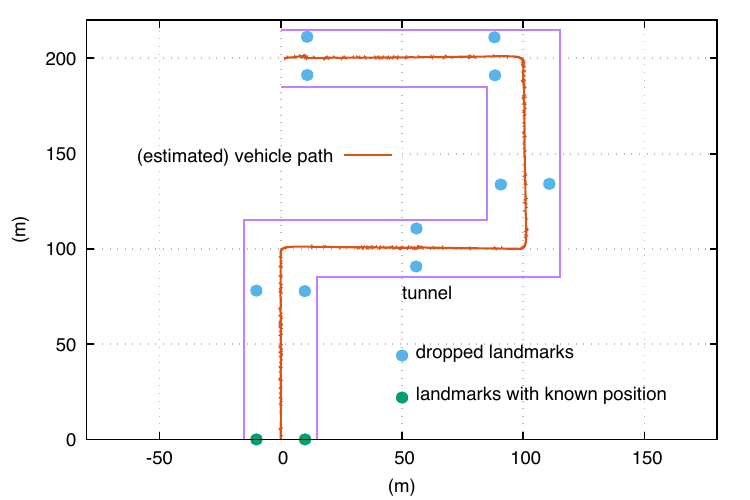}
    \caption{Landmark drop locations computed by the DNN for straight tunnel with length $400$  \si{\meter} applied to a tunnel with $3$ turns. }
    \label{fig:lm_st_with_turns}
\end{figure}

\begin{figure}
    \centering
    \includegraphics{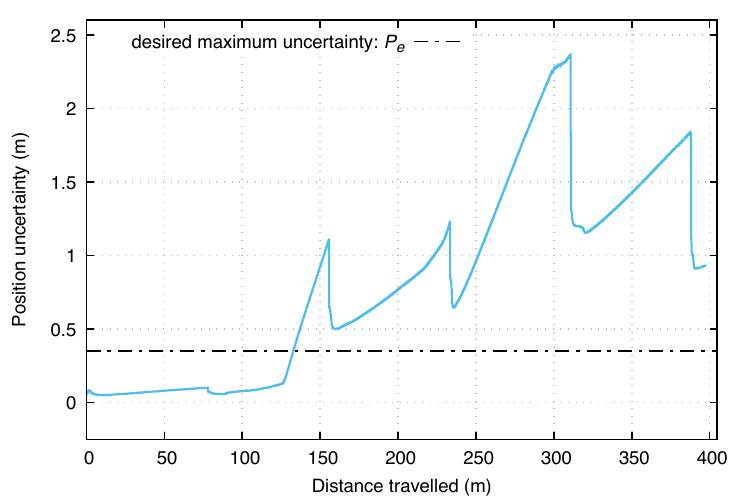}
    \caption{Instantaneous position uncertainty plot for landmark drop locations computed by the DNN for straight tunnel with length $400$ \si{\meter} when applied to a tunnel with $3$ turns. The instantaneous position uncertainty exceeds the limit $P_e$ in this case, motivating the need for landmark drop-adjustment algorithms.}
    \label{fig:place_st_with_turns}
\end{figure}

\begin{figure}
    \centering
    \includegraphics{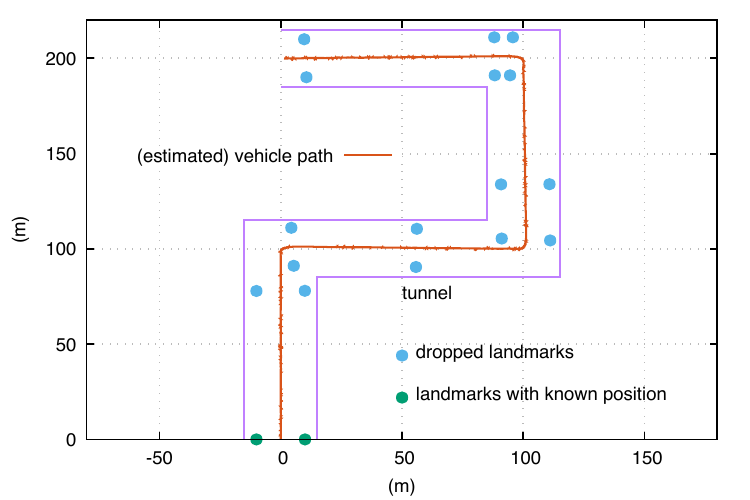}
    \caption{Landmark drop locations computed by the drop adjustment algorithm given the information on the length of the tunnel and the number of turns.}
    \label{fig:lm_con_with_turns}
\end{figure}

\begin{figure}
    \centering
    \includegraphics{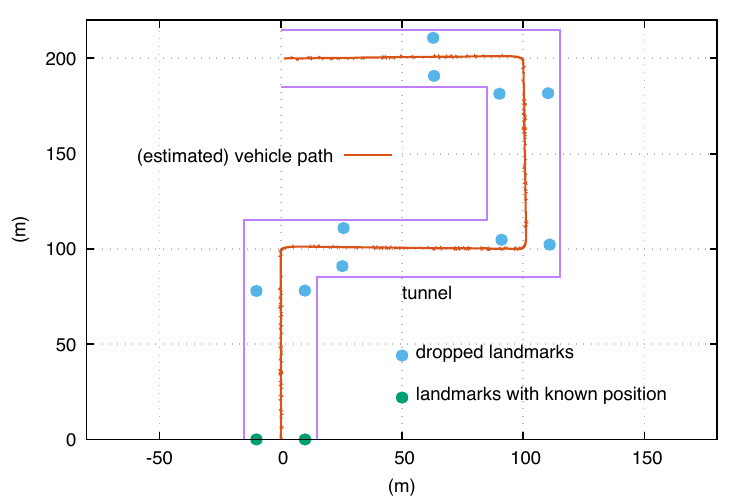}
    \caption{Landmark drop locations computed by the drop adjustment algorithm given the full topology of the tunnel.}
    \label{fig:lm_op_with_turns}
\end{figure}

We now consider a tunnel with $3$ turns (see Fig. \ref{fig:lm_st_with_turns} -- \ref{fig:lm_op_with_turns}). The length of the tunnel is $400$ \si{\meter}. The other parameters for this experiment are as follows: $v= 3$ \si{\meter\per\second}, $\sigma_v = 0.3$ \si{\meter\per\second}, $\rho^{\max} = 90$ \si{\meter} and $\sigma_{\ell} = 0.1$ \si{\meter}. Then the value of the drop distance for $P_e = 0.35$ is obtained by solving the MILP. The drop locations of the landmarks for drop distance obtained using the MILP is shown in Fig. \ref{fig:lm_st_with_turns}. The turns in the tunnel will cause the range measurements from the landmark drop due to loss of LOS. This results in the instantaneous position uncertainty exceeding the user-specified maximum position uncertainty limit $P_e$ (see Fig. \ref{fig:place_st_with_turns}). This also motivates the need for the landmark drop-adjustment algorithms in Sec. \ref{sec:drop-adjustment}. 

\begin{figure}
    \centering
    \includegraphics{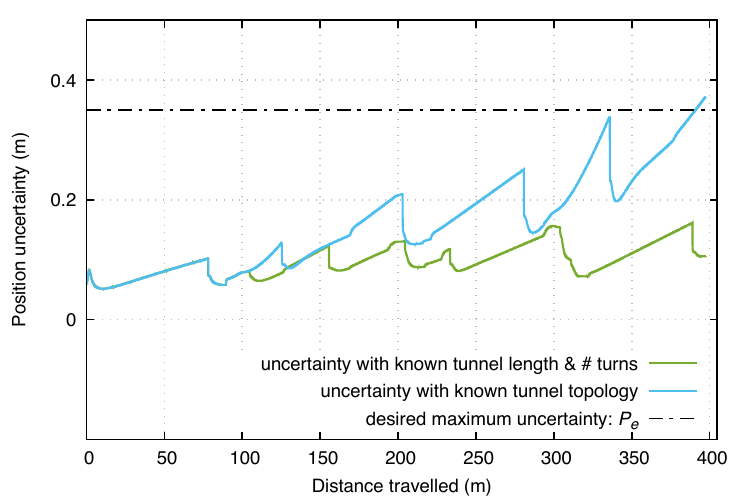}
    \caption{Instantaneous position uncertainty plot for landmark drop locations provided by the landmark drop-adjustment algorithms for length $400$ \si{\meter} with 3 turns.}
    \label{fig:place_with_info}
\end{figure}

The landmark drop locations computed using the DNN are adjusted using the two algorithms in Sec. \ref{sec:drop-adjustment} when the number of turns and the full topology of the tunnel are known, respectively. The landmark drops for the tunnel with the two drop adjustment algorithms are shown in Fig. \ref{fig:lm_con_with_turns} and \ref{fig:lm_op_with_turns}. The respective instantaneous position uncertainties are shown in Fig. \ref{fig:place_with_info}. As observed in Fig. \ref{fig:place_with_info}, the maximum position uncertainty value for the entire path is approximately close to the desired value indicating the effectiveness of the drop-adjustment algorithms in computing drop locations while ensuring mission constraints are satisfied. 

\subsection{Effect of staggering landmark drop locations} \label{subsec:staggering}
\begin{figure}
    \centering
    \includegraphics{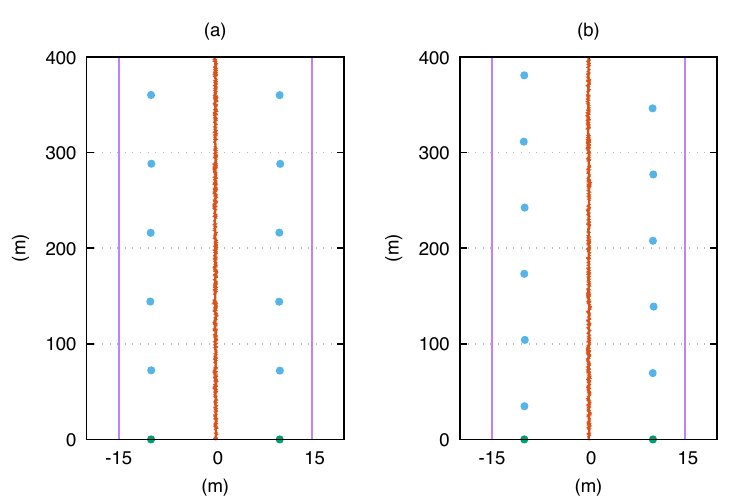}
    \caption{Placement of landmarks and estimated vehicle path when (a) a pair of landmarks are placed at either side of the vehicle's path at a drop distance of $d^*$ \si{\meter} (b) one landmark is placed every $d^*/2$ \si{\meter}. For these plots, $d^* = 72$ \si{\meter}. The green circles are the known landmarks, the blue circles are dropped landmarks and the red line is the estimated vehicle's path. The purple lines denote the tunnel walls.}
    \label{fig:staggered-straight-loc}
\end{figure}

\begin{figure}
    \centering
    \includegraphics{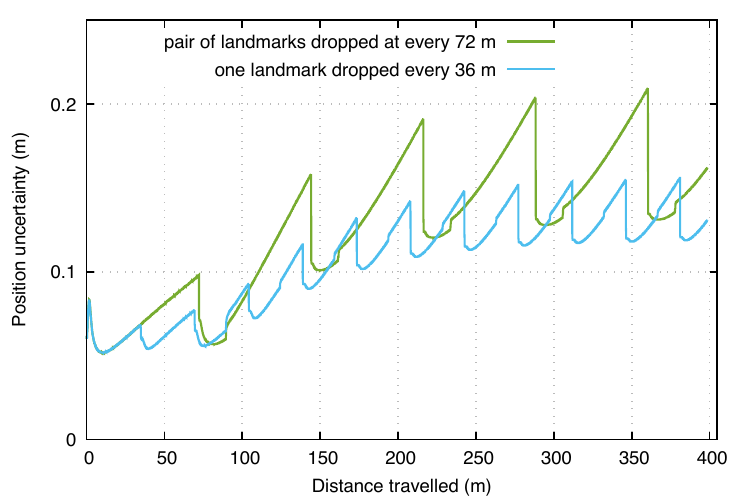}
    \caption{Improvement in position uncertainty when landmark drops are staggered at half the drop distance.}
    \label{fig:staggered-straight-pos}
\end{figure}

\review{In Sec. \ref{sec:statement}, it was assumed that at each landmark drop location, a pair of landmarks, one to the left and one to the right of the vehicle are dropped. This was mainly done to ensure that the vehicle always has measurements from at least two distinct landmarks. In this section, we present some alternatives of staggering the landmark drops and examine the effect it has on the position uncertainty. We first present the effect of staggering the landmark drops on a straight tunnel. The staggering of landmark drops for the straight tunnel is performed as follows: if the drop distance obtained by solving the MILP for a straight tunnel is given by $d^*$, then one landmark is dropped at every $d^*/2$ units. The drops are alternated between the left and right of the vehicle. For a drop distance of $72$ \si{\meter}, the landmark drop locations and the corresponding position uncertainties for the staggered and non-staggered drops are shown in Fig. \ref{fig:staggered-straight-loc} and \ref{fig:staggered-straight-pos}, respectively. }


\begin{figure}
    \centering
    \includegraphics{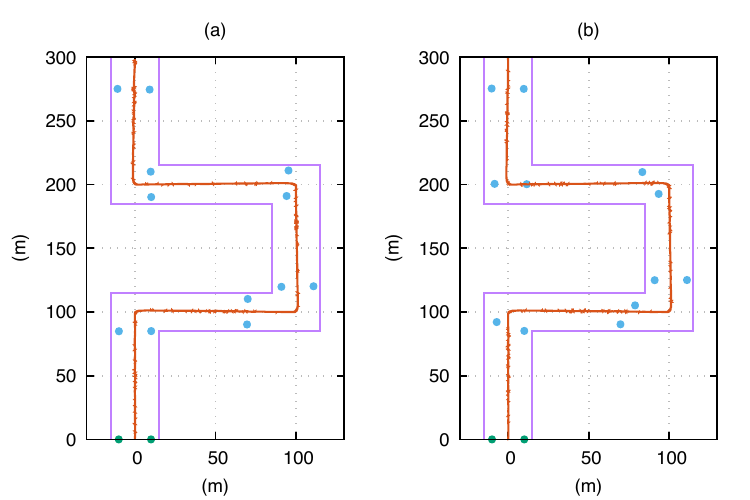}
    \caption{Placement of landmarks and estimated vehicle path when (a) a pair of landmarks are placed at either side of the vehicle's path (b) landmarks are staggered at an angle to each other near a tunnel turn. The green circles are the known landmarks, the blue circles are dropped landmarks and the red line is the estimated vehicle's path. The purple lines denote the tunnel walls.}
    \label{fig:staggered-turn}
\end{figure}

\begin{figure}
    \centering
    \includegraphics{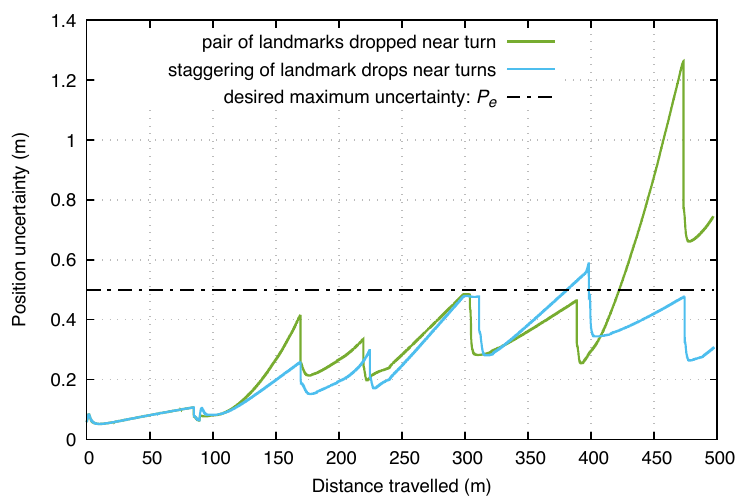}
    \caption{Improvement in position uncertainty when landmark drops are staggered near the turns.}
    \label{fig:staggered-pos}
\end{figure}

The other instance where the staggering the landmark drops can be effective is when landmark drop points are close to a turn. When drop locations are close to the turn, staggering can also aid in tackling the the effective co-linearity issue shown in Fig. \ref{fig:l-obs}. In this case when drop locations are close to a turn, the two landmarks are dropped slightly at an angle to each other and this aids in neutralizing the co-linearity issue. The landmark drops and the corresponding position uncertainty are shown in Fig. \ref{fig:staggered-turn} and \ref{fig:staggered-pos}, respectively. Later in the next section, we show that wall-update sensors can also address the co-linearity issue that is tackled by staggering the landmark drops near turns. 

\subsection{Advantages of wall-update sensors} 
\begin{figure}
    \centering
    \includegraphics{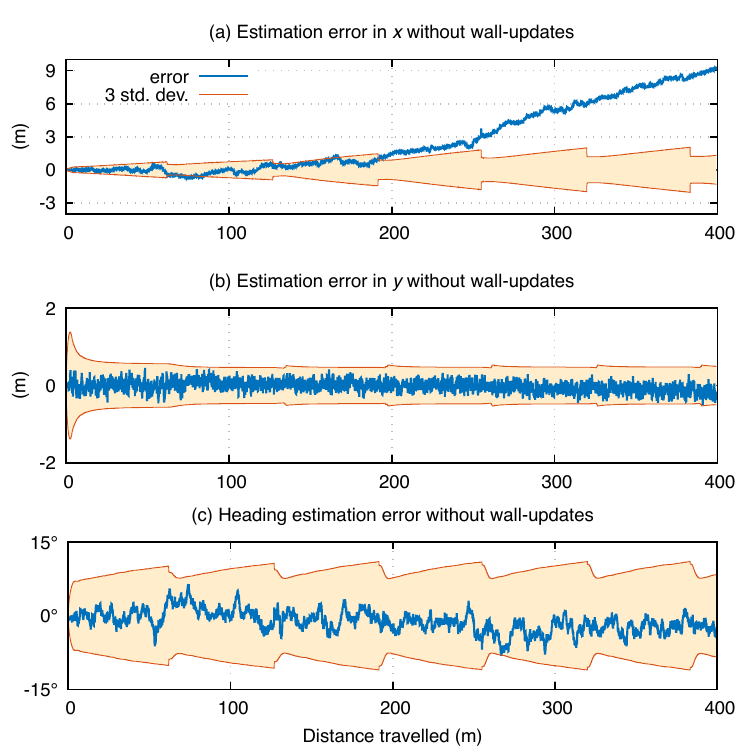}
    \caption{Position and heading errors when vehicle has no wall-update sensors.}
    \label{fig:no_wall_updates}
\end{figure}

\begin{figure}
    \centering
    \includegraphics{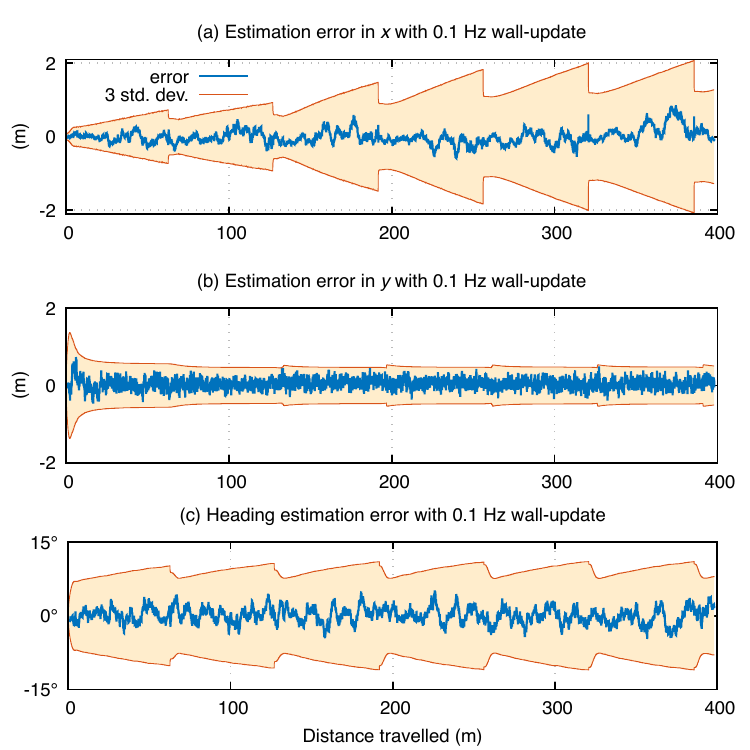}
    \caption{Position and heading errors when vehicle utilizes measurements from wall-update sensors at $0.1$ \si{\hertz}.}
    \label{fig:wall_01_updates}
\end{figure}

\begin{figure}
    \centering
    \includegraphics{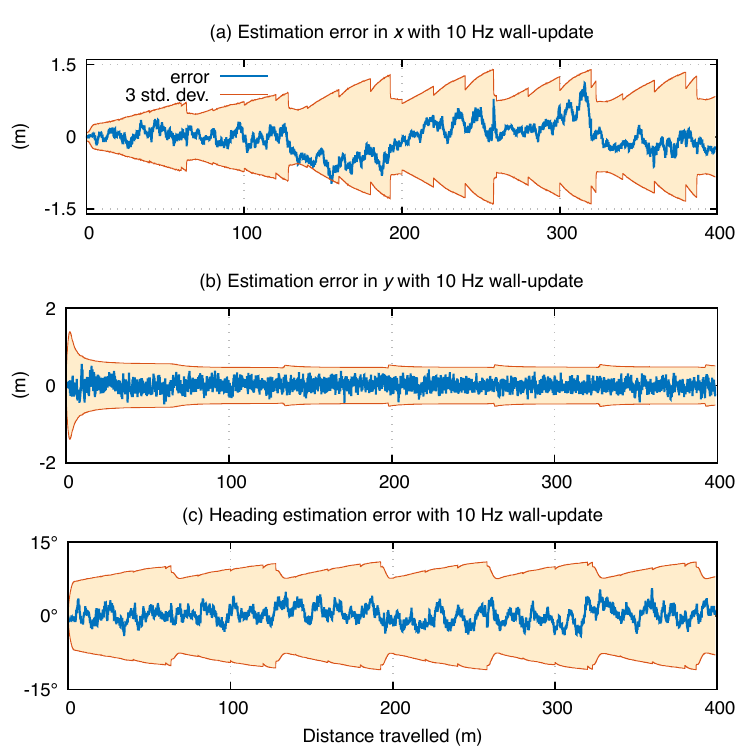}
    \caption{Position and heading errors when vehicle utilizes measurements from wall-update sensors at $10$ \si{\hertz}.}
    \label{fig:wall_10_updates}
\end{figure}

\begin{figure}
    \centering
    \includegraphics{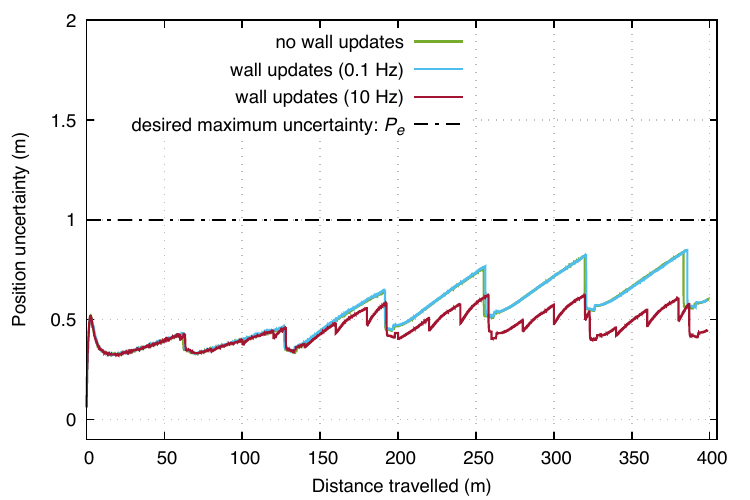}
    \caption{Position uncertainty when the vehicle navigates through the straight tunnel with and without wall-update sensors}
    \label{fig:wall_update_uncertainty}
\end{figure}

This section is aimed at showing the efficacy of the wall-update sensors in (a) preventing bias in the EKF algorithm and (b) ensuring the position uncertainty does not increase erratically if two landmarks (from which the vehicle obtains range measurements) and the vehicle itself are \textit{effectively} co-linear. To show the bias-prevention properties of the wall-update sensors, we consider a straight tunnel with length $400$ \si{meter}. The other parameters for this experiment are as follows: $P_e = 2.5$, \si{\meter}, $v= 2$ \si{\meter\per\second}, $\sigma_v = 0.5$ \si{\meter\per\second}, $\rho^{\max} = 135$ \si{\meter}, and $\sigma_{\ell} = 1$ \si{\meter}. 
The drop distance for the above setup is obtained by solving the MILP and the EKF is run with the vehicle dropping two landmarks periodically with a period equal to drop distance. With this setup, we run the simulation for three cases: (i) without any wall-update sensors (ii) wall-update sensors operating at a frequency of $0.1$ \si{\hertz} and (iii) wall-update sensors operating at a frequency of $10$ \si{\hertz}. For all three cases, the estimation error and associated uncertainty (3 std. dev. bounds) for the vehicle's position and heading are shown in Fig. \ref{fig:no_wall_updates} -- \ref{fig:wall_10_updates}. As observed from Fig. \ref{fig:no_wall_updates} and \ref{fig:wall_update_uncertainty}, the EKF has high confidence in the position and heading estimates as depicted by the $3\sigma$ bounds and the position uncertainty in the plots. Nevertheless, the error in the $x$-position of the estimate (see Fig. \ref{fig:no_wall_updates} (a)) keeps increasing. This is the bias in the $x$-position estimate of the vehicle. As seen from the plots Fig. \ref{fig:wall_01_updates} and \ref{fig:wall_10_updates}, this bias is prevented when wall-update sensors are utilized. Furthermore, increasing the update frequency of the wall-update sensors improves the performance of the overall navigation system considerably. 

\begin{figure}
    \centering
    \includegraphics{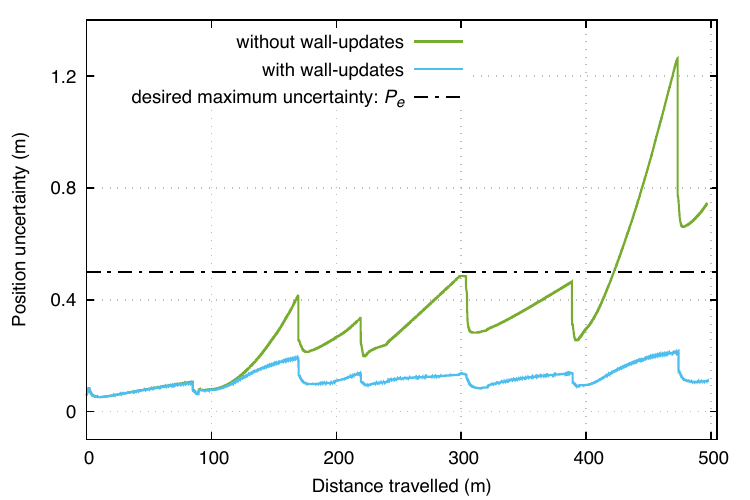}
    \caption{Position uncertainty of the vehicle when the two landmarks and the vehicle are effectively co-linear.}
    \label{fig:wall_1}
\end{figure}

The Fig. \ref{fig:wall_1} presents the position uncertainty for a tunnel with length $500$ \si{\meter} and $3$ turns. The landmark drop locations were chosen by solving the MILP followed by the landmark drop-adjustment algorithm, assuming we have the complete topology of the tunnel. The value of $P_e$ was set to $0.5$. The sensing noise parameters remain the same as in the previous study. Effective co-linearity results in the position uncertainty increasing beyond the user-specified limit of $P_e = 0.5$ as shown in Fig. \ref{fig:wall_1}. This increase does not occur when wall-update sensors are added to the navigation system, corroborating its effectiveness. 

\section{Practical considerations and future work} \label{sec:practical}
\review{In this section, we briefly present (i) challenges that may arise during hardware implementation of the proposed algorithms using actual drones and (ii) in doing so, identifies avenues for future work. Nevertheless, wherever possible, we also present simple solution techniques that can be added to the proposed system to tackle some of the challenges. }

\review{In practice, the localization payload a single UV can carry would depend upon the dimension, weight, sensing range and noise of the RFID tags. Since the quality of these sensors are improving, the robustness of our approach is subjected to the advancement of the quality of these tags \cite{Decawave1001}. Using low quality sensors in combination with vehicles with less payload capacity in the proposed navigation system severely restricts the exploration distance for the mission. This challenge can be addressed by using a multi-UV system, each carrying a set of localization payload and benefiting from the payloads that are deployed from each of them. In this context, future work would focus on developing navigation systems with multiple UVs with shared payload deployment capability. These multi-UV systems will especially be valuable in the case where only the tunnel's length and the number of turns are known (see Sec. \ref{subsec:known-turns}).} 

\review{The second challenge arises due to an assumption in our proposed proposed navigation system that all the landmarks are dropped on the floor on the either side of the vehicle's path. This assumption completely ignores ground effects and its impact on the quality of localization. The challenge can be tacked by developing mechanisms that can deploy these landmarks directly on to the sides of the tunnel or on the ceiling. Future work would focus on explicitly modeling these ground effects and their impact on the localization algorithm. Furthermore, the wall-update sensors in Sec. \ref{subsec:wall-update} assume that the walls on the either side of the tunnels are fairly smooth that there are no obstacles and other objects in the tunnel. Future work would also focus on making the proposed algorithms robust to external obstacles or objects in the environment that the have a deteriorating effect on the measurements. Finally, algorithms that can stagger the landmark drop locations in a systematic way is yet another avenue for future work.}

\section{Conclusion} \label{sec:conclusion}
\review{The paper presents the first data-driven, deployable navigation system for UVs in GPS-denied, feature-deficient environments like tunnels and mines. To the best of our knowledge, this is the first work that combines techniques from machine learning, estimation and mathematical programming to develop such a navigation system. The effectiveness of both the proposed navigation system and its individual components is corroborated through extensive simulation experiments on environments with different configurations. The other main takeaway from this paper that unobservability is not necessarily a bad thing from an localization stand-point and by combining techniques from disparate fields, the rate of increase of localization errors can be sufficiently controlled to build good vehicle navigation systems in GPS-denied, feature-deficient environments. Finally, The system stands out in the sense that it does not require heavy computing, capabilities unlike SLAM. Nevertheless, this system is not intended to replace SLAM; rather, it is seen as an alternative to SLAM in feature-deficient small environments. }

\section{Declarations}
\begin{itemize}
\item \textbf{Funding}: The authors acknowledge Air Force Research Laboratory, Grant FA8651-16-1-0001 and LANL's Lab Directed Research and Development program (LDRD) project 20200016DR for funding this work.
\item \textbf{Conflicts of interest}: None
\item \textbf{Code availability}: Not available publicly. 
\item \textbf{Authors' Contributions}: The following author's worked on this manuscript.
\begin{enumerate}
\item Sohum Misra
\item Kaarthik Sundar
\item Rajnikant Sharma
\item Kevin Brink
\end{enumerate}

Sohum Misra and Kaarthik Sundar conceived the idea of combining learning and MILP and the presented idea. Sohum Misra performed the analytic calculations and performed the numerical
simulations. Rajnikant Sharma and
Kevin Brink supervised the findings of this work. All authors discussed the results and contributed to the final manuscript.

\item \textbf{Ethics approval}: Not applicable
\item \textbf{Consent to participate}: Not applicable 
\item \textbf{Consent for publication}: All the authors listed above have agreed to publish this work.
\end{itemize}
\printbibliography

\end{document}